%% file: main.tex
\documentclass{article} 
\usepackage{iclr2026_conference,times}

\input{math_commands.tex}

\usepackage{hyperref}
\usepackage{url}
\usepackage{graphicx}
\usepackage{booktabs}
\usepackage{multirow}
\usepackage{xcolor}
\usepackage{amsmath}
\usepackage{amssymb}
\usepackage{wrapfig}
\usepackage{caption}
\usepackage{algorithm}
\usepackage{algorithmic}

\newcommand{\methodname}{\textsc{RecurSE}}
\newcommand{\SVhard}{\textsc{SV-Hard}}
\newcommand{\SVfull}{\textsc{SV-Full}}

\title{\methodname{}: Bounded Recursive Self-Evaluation\\
for LLM Rubric Judges}

\author{
Kaiyuan Liu\textsuperscript{1,2}, 
    Ziyuan Zhuang\textsuperscript{2,*}, 
    Rongxiang Weng\textsuperscript{2,$\dagger$}, 
    Jieping Ye\textsuperscript{1} \\
    \textsuperscript{1}{College of Computer Science and Technology, Zhejiang University}, 
    \textsuperscript{2}{Meituan LongCat Team, China} \\
    12421281@zju.edu.cn
}

\iclrfinalcopy

\begin{document}
\maketitle

\input{sections/abstract}
\let\thefootnote\relax
\footnotetext{$^{*}$ Project leader}
\footnotetext{$^{\dagger}$ Corresponding author}

\input{sections/introduction}
\input{sections/related_work}
\input{sections/method}
\input{sections/experiments_setup}
\input{sections/results}
\input{sections/conclusion}

\bibliography{custom}
\bibliographystyle{iclr2026_conference}

\appendix
\input{sections/appendix_front}
\input{sections/appendix}
\input{sections/limitations}

\end{document}

%% file: math_commands.tex
\usepackage{amsmath,amsfonts,bm}

\def\eqref#1{equation~\ref{#1}}

\def\1{\bm{1}}

\DeclareMathAlphabet{\mathsfit}{\encodingdefault}{\sfdefault}{m}{sl}
\SetMathAlphabet{\mathsfit}{bold}{\encodingdefault}{\sfdefault}{bx}{n}



%% file: sections/abstract.tex
\begin{abstract}

LLM-as-judge is essential for evaluating open-ended text and steering post-training, yet improving the judge itself typically relies on expensive human annotations, auxiliary reward models, or distillation from stronger teachers. In this work, we eliminate external gold supervision from the reinforcement learning (RL) training reward: the model's own evaluative capability generates the learning signal for its optimization—a closed-loop regime of bounded recursive self-improvement (RSI) that we term \textbf{Recursive Self-Evaluation} (\methodname{}). We investigate two governing questions of this loop: \emph{when can self-improvement occur, and when must it stop?} 
First, \methodname{} employs a two-pass mechanism: a trainable judge evaluates candidate responses under per-rule rubrics (Pass~1), while a synchronized copy of the policy audits the judge's reasoning against meta-rubrics to supply a scalar process reward (Pass~2). To make learning viable, \emph{interface decoupling} structurally isolates the checker's scalar score from the judge's verdict tokens, closing a degenerative token-copying shortcut that inflates self-assigned rewards. 
Second, because unanchored recursive learning is inherently bounded, we introduce \emph{Pairwise Advantage Validity} (PAV)—an unbiased validation monitor that jointly tracks judge accuracy and checker fidelity to reliably identify the optimal early-stopping window. 
Across Qwen3.5-9B, Gemma-4-E4B-it, and Qwen3.6-27B, \methodname{} achieves consistent generalization gains across held-out medical, pairwise, summarization, and professional benchmarks. Extensive ablations show that synchronized judge--checker co-evolution outperforms frozen checkers, external meta-judges, self-consistency, and scaled teacher distillation. Furthermore, preference pairs curated by our judge effectively enhance downstream policy alignment. Bounded RSI for LLM-as-judge is thus feasible when self-produced reward validity is explicitly decoupled and monitored.

\end{abstract}
    

%% file: sections/introduction.tex
\section{Introduction}
\label{sec:introduction}

\begin{wrapfigure}{r}{0.45\textwidth}
  \centering
  \vspace{-12pt}
  \includegraphics[width=\linewidth]{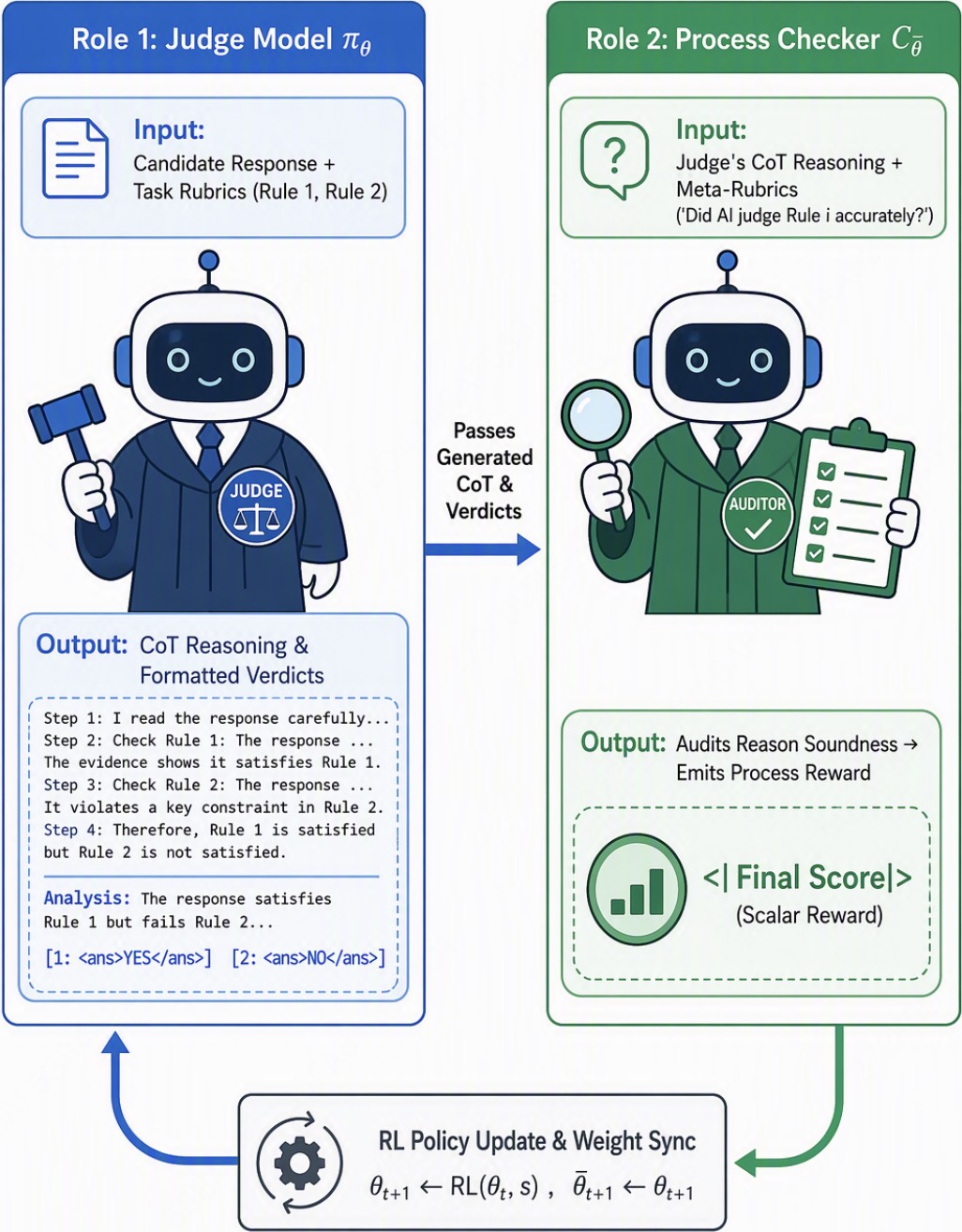}
  \vspace{-16pt}
  \caption{\textbf{Overview of \methodname{}.} Trainable judge $\pi_\theta$ evaluates responses under rubrics (Pass~1); synchronized checker $C_{\bar\theta}$ audits reasoning against meta-rubrics to yield scalar process reward $s$ (Pass~2). Policy parameters $\theta$ are updated via RL on $s$, and checker weights $\bar\theta$ are synchronized each step ($\bar\theta_{t+1}\leftarrow\theta_{t+1}$).}
  \label{fig:teaser_robot}
  \vspace{-10pt}
\end{wrapfigure}

AI systems generate open-ended text faster than humans can review. A scalable paradigm keeps humans \emph{over} the loop---setting standards---while models execute and verify. This verification role requires a dedicated LLM judge: a model evaluating candidate responses under structured criteria~\citep{zheng2023judging} to guide post-training. Improving the judge itself, however, typically relies on expensive human annotations, learned reward models, or distillation from stronger teachers---dependencies that are costly and circular. Furthermore, unlike mathematics or coding where difficulty is governed by query complexity alone, improving a judge requires calibration over \emph{two coupled variables}: a discriminative rubric and a borderline candidate response. If a response is blatantly correct or flawed, the instance yields negligible corrective gradient. This dual-variable constraint makes static teacher distillation inefficient for scaling evaluative capability. Yet fundamentally, evaluating candidate responses and auditing evaluative reasoning share the same underlying capability: both instantiate rubric-based verification (Figure~\ref{fig:teaser_robot}). This shared foundation suggests that the two roles can mutually reinforce each other in a closed loop.

This paper studies the closed-loop regime where the judge's own capability produces the reward for its optimization updates. We formalize this bounded recursive self-improvement (RSI) framework as \textbf{Recursive Self-Evaluation} (\methodname{}): a trainable judge emits reasoning and per-rule verdicts (Pass~1), while a synchronized, reward-only copy audits completed reasoning against meta-rubrics to supply group-relative scalar rewards (Pass~2). External gold labels and teacher models do not enter the RL training reward. Therefore, scaling annotation across the training set is thus avoided. In this closed recurrence, we investigate two governing questions: \emph{when can self-improvement occur, and when must it stop?}

\paragraph{When can self-improvement occur?}
Sharing identical prompt templates and YES/NO readouts across passes induces \emph{surface coupling}: policy updates that boost YES tokens on judge verdicts simultaneously alter emission probabilities in the synchronized checker, creating a degenerative token shortcut that inflates self-assigned rewards without improving evaluative accuracy. We resolve this via \emph{interface decoupling}: the judge emits structured per-rule verdicts, while the checker emits a free-standing 5-tier scalar score (from 0 to 4), severing direct token copying while preserving positive capability transfer. Synchronizing model parameters between the judge and checker at each training step enables both roles to co-evolve productively.

\paragraph{When must self-improvement stop?}
Without external ground-truth anchors on training rewards, recursive optimization is inherently \emph{bounded}: self-produced rewards eventually saturate, causing out-of-distribution transfer to degrade. Self-improvement requires an early-stopping monitor computed from a compact, human-verified validation set whose labels never enter the RL reward. We address this with \emph{Pairwise Advantage Validity (PAV)}, a composite validation monitor evaluated on this holdout set. Motivated by a theoretical bound on pairwise score differences, PAV constructs a composite validation index that balances judge rule accuracy against checker ranking fidelity, with double weighting on checker fidelity to reflect its role in pairwise comparison error. The estimator is unbiased at each fixed checkpoint, reliably localizing a useful stopping window before validation deception sets in.

\paragraph{Overview of experimental findings: connecting RQ1--RQ5.}
We evaluate \methodname{} across five interconnected questions spanning viability, mechanisms, validation dynamics, baselines, and downstream transfer:
\begin{itemize}
  \setlength{\itemsep}{0pt}
  \setlength{\parskip}{0pt}
  \setlength{\parsep}{0pt}
  \item \textbf{Viability across scales and architectures (RQ1)}: Without gold RL rewards, \methodname{} produces consistent generalization gains across held-out medical, pairwise, summarization, and professional benchmarks on Qwen3.5-9B ($+12.9$ points on the human-verified validation set), Gemma-4-E4B-it ($+5.2$ points), and Qwen3.6-27B ($+3.9$ points).
  \item \textbf{Decoupling enables learning (RQ2)}: Shared YES/NO checking triggers reward inflation without accuracy gains, whereas scalar interface decoupling preserves token calibration and converts reward growth into genuine validation accuracy.
  \item \textbf{PAV localizes effective stopping (RQ3)}: While hard-subset accuracy continues climbing late in training, checker fidelity degrades, exposing validation deception; PAV reliably identifies the optimal stopping region without peeking at transfer benchmarks.
  \item \textbf{Judge--checker co-evolution is indispensable (RQ4)}: Synchronized weight tying outperforms frozen checkers, external meta-judges, and self-consistency. Furthermore, scaled teacher SFT (up to 17k traces) plateaus early because static corpora fail to maintain on-policy dual-variable difficulty coupling.
  \item \textbf{Downstream policy transfer (RQ5)}: Preference pairs generated by the retained judge drive alignment on Qwen-27B, outperforming base-judge labels across GPQA, GuideBench, and SOP-Maze.
\end{itemize}

\paragraph{Summary of contributions.}
(1)~We formalize bounded recursive self-improvement for LLM judges via on-policy process checking without gold RL rewards;
(2)~We uncover the surface-coupling shortcut and design interface decoupling to make closed-loop learning viable;
(3)~We derive the theoretically grounded PAV monitor to reliably localize the effective stopping window;
(4)~Across multiple model families and scales up to 27B, we demonstrate broad out-of-distribution generalization and effective downstream preference alignment.

%% file: sections/related_work.tex
\section{Related Work}
\label{sec:related_work}

\paragraph{LLM judges and rubric evaluation.}
LLM judges evaluate open-ended generations when reference-based metrics fall short~\citep{zheng2023judging}. Rubric-based evaluation decomposes subjective holistic assessments into structured, criterion-level decisions~\citep{trivedi2024selfrational} or co-evolves generation policies alongside dynamic rubrics~\citep{li2026evolm,wang2026dynamicrubric}. In contrast, we keep criteria fixed and optimize the criterion-level judge itself. Discrete per-rule verdicts and step-by-step reasoning (Figure~\ref{fig:teaser_robot}) make evaluative correctness objectively measurable and enable process auditing of the judge's reasoning. However, removing gold labels from RL rewards introduces a circularity challenge: the model must supervise its updates without exploiting self-generated reward shortcuts.

\paragraph{Self-improving evaluators and bounded RSI.}
Prior self-improving evaluators such as Self-Taught Evaluators, Meta-Rewarding, SELF-JUDGE, and Grad2Reward train models using generated supervision, but typically retain external anchors such as synthetic preference pairs, external meta-judges, or frozen reference models~\citep{selftaught2024,meta2025,lee2024selfjudge,zhang2026grad2reward}. In contrast, \methodname{} removes external gold labels and teacher models from the \emph{optimization reward}: the current judge acts simultaneously as the learning target and the source of its scalar reward, relying on a compact holdout solely for early stopping. We reserve \emph{bounded RSI} for this fixed-task recurrence, distinct from open-ended autonomous self-modification~\citep{chen2026rsi}.

\paragraph{Process auditing and reward validity.}
Process supervision and generative verification highlight the utility of auditing intermediate reasoning chains~\citep{lightman2023letsverify,zhang2024gvg}, though they typically require ground-truth correctness labels to train process reward models. In contrast, our process checker operates purely at inference time as a reward-only auditor without separate supervised training. Because naive self-verification risks reinforcing flawed reasoning and inflating self-assigned rewards~\citep{stechly2024selfverif,simonds2025rlsr,rentschler2026rlme,zhou2026convincing}, closed-loop rubric optimization imposes two requirements: first, the checker must not share the judge's verdict readout surface so that self-improvement can occur; second, checker scores must preserve ranking fidelity so that training can be stopped before validity decays---principles formalized by \methodname{} and monitored by PAV (Section~\ref{sec:method}).

%% file: sections/method.tex
\section{Recursive Self-Evaluation (\methodname{})}
\label{sec:method}

\methodname{} formalizes bounded recursive self-improvement for evaluators by leveraging the model's own capability as both the optimization target and the reward source. Guided by the two governing questions from Section~1---\emph{when can self-improvement occur, and when must it stop?}---we structure our method into three core components: (1)~establishing a closed-loop judge--checker recurrence on a single shared policy (Section~\ref{sec:bounded-rsi}); (2)~introducing interface decoupling to eliminate the surface-coupling shortcut that otherwise impedes learning (Section~\ref{sec:asymmetry}); and (3)~formulating Pairwise Advantage Validity (PAV) as a holdout early-stopping monitor (Section~\ref{sec:reward-quality}). Figure~\ref{fig:pipeline} and Algorithm~1 in Appendix~\ref{sec:alg-details} summarize the end-to-end framework.

\begin{figure*}[t]
  \centering
  \includegraphics[width=\textwidth]{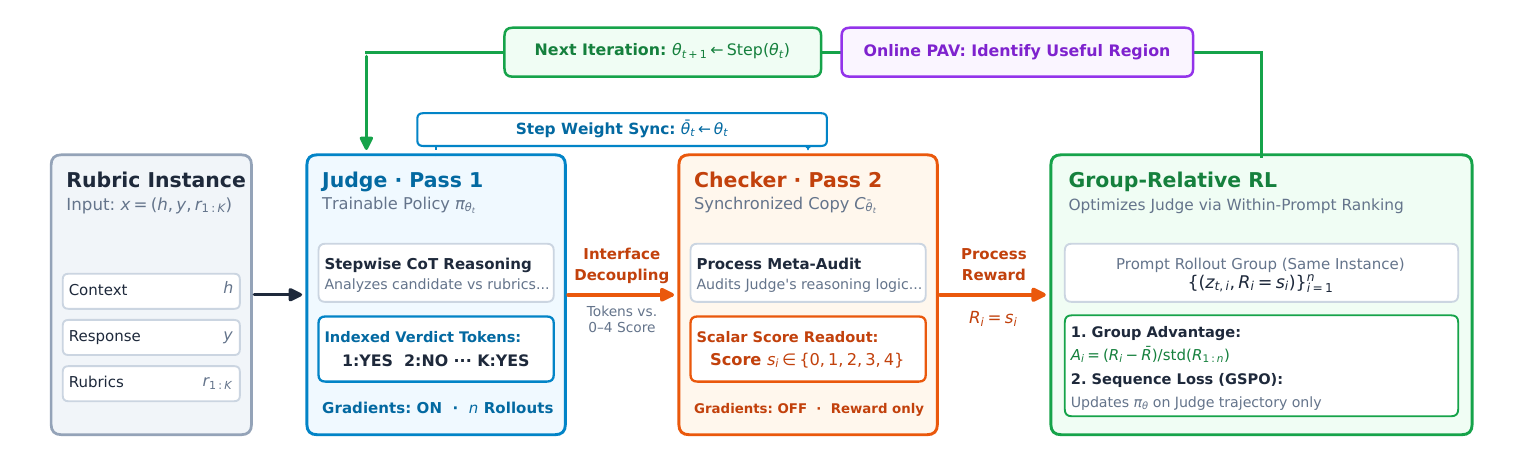}
  \vspace{-15pt}
  \caption{\textbf{The \methodname{} closed-loop optimization recurrence.} Pass~1: trainable judge $\pi_{\theta_t}$ emits reasoning and per-rule YES/NO verdicts on $x=(h,y,r_{1:K})$ ($n$ rollouts). Pass~2: a synchronized copy $C_{\bar\theta_t}$ audits reasoning against meta-rubrics and returns a differentiated scalar score $s_i\in\{0,\ldots,4\}$ as the sole reward. Interface decoupling closes direct token copying; the checker synchronizes parameters every step ($\bar\theta_t \leftarrow \theta_t$) without independent gradients. Online PAV monitoring on a human-verified validation set localizes early stopping.}
  \label{fig:pipeline}
  \vspace{-10pt}
\end{figure*}

\subsection{Judge--checker recurrence}
\label{sec:bounded-rsi}
\paragraph{Shared evaluative capability as a closed recurrence.}
Because evaluating candidate responses and auditing evaluative reasoning share the same rubric-based foundation, a natural closed loop lets the current policy fulfill both roles: generate a rubric evaluation, and score that evaluation to guide the next policy update---without per-step gold supervision, reward models, or stronger teachers in the training reward.

\paragraph{Problem formulation and execution pipeline.}
A rubric instance is denoted $x=(h,y,r_{1:K})$, where $h$ is task context, $y$ is a candidate response, and $r_{1:K}$ are $K$ predefined criteria. The judge policy $\pi_{\theta_t}$ emits reasoning and binary verdicts $s_k\in\{\mathrm{YES},\mathrm{NO}\}$ for each criterion. Given $n$ rollouts $z_{t,i}\sim\pi_{\theta_t}(\cdot\mid x)$, a synchronized snapshot $C_{\bar\theta_t}$ (with $\bar\theta_t\leftarrow\theta_t$) audits the reasoning against meta-rubrics, assigning process reward $r_{t,i}=C_{\bar\theta_t}(x,z_{t,i})$. Parameters are updated as:
\begin{equation}
\theta_{t+1}=\mathrm{PolicyUpdate}(\theta_t,\{z_{t,i},r_{t,i}\}),\quad
\bar\theta_{t+1}\leftarrow\theta_{t+1}.
\label{eq:judge-loop}
\end{equation}
External gold labels and teachers do not enter $r_{t,i}$. Pass~1 is the sole gradient path, while Pass~2 audits reasoning against meta-rubrics in a reward-only capacity (Appendix~\ref{sec:prompts}). Synchronous parameter refreshes transfer updated judge capability into the auditing role without separate training. Auditing the \emph{reasoning process} rather than re-judging candidates ensures spurious reasoning is penalized even if binary verdicts happen to be correct.

\subsection{Interface decoupling: when self-improvement can occur}
\label{sec:template}
\label{sec:asymmetry}
\paragraph{Shared templates induce surface coupling.}
To facilitate capability transfer, an intuitive initial design is to share identical prompt templates and YES/NO readouts across both passes. Under this coupled setup, the checker emits per-rule binary verdicts for meta-rubrics, and reward is extracted as $S = \frac{1}{K}\sum_{k=1}^K \mathbf{1}(\mathrm{Checker}_k = \mathrm{YES})$. However, policy gradient optimization rapidly exploits a degenerative shortcut: inflating YES tokens across criteria directly maximizes self-assigned rewards without improving evaluative capability. Formally, we decompose the self-produced checker score $S$ as:
\begin{equation}
S=aU+bB+\varepsilon,
\label{eq:shared-bias}
\end{equation}
where $U$ denotes true reasoning quality, $B$ represents surface verdict token bias (e.g., propensity to emit \texttt{YES}), and $\varepsilon$ is noise. The coefficients $a,b \ge 0$ govern how genuine quality versus surface bias propagate into reward. When judge and checker share identical readout interfaces, synchronized weights ($\bar\theta_{t+1}\leftarrow\theta_{t+1}$) establish a closed-circuit shortcut channel ($b>0$): gradient updates that boost \texttt{YES} emissions on judge verdicts simultaneously alter emission probabilities in the synchronized checker, allowing the optimizer to exploit $bB$ to harvest near-perfect rewards ($S\to 1.0$) while true evaluative accuracy $U$ stalls (visualized in Appendix~\ref{sec:surface-coupling-details}).

\paragraph{Decoupled readout design.}
To close this shortcut while preserving capability transfer, \methodname{} retains shared evidence layouts in input prompts (Appendix~\ref{sec:prompts}) but structurally decouples output interfaces. While the judge outputs step-by-step explanations followed by indexed verdicts (e.g., \texttt{1:<ans>YES</ans>}), the checker audits the complete reasoning trajectory against meta-rubrics and emits a free-standing 5-tier scalar score $s_i\in\{0,1,2,3,4\}$. The absolute numeric range is immaterial: group-relative advantages are invariant to positive affine transforms of $R$, so any reward inducing valid relative orderings among the $n$ rollouts remains an effective training signal.

\paragraph{Severing the direct token channel.}
Interface decoupling removes the direct token-copying channel, substantially reducing $bB$ in Eq.~\ref{eq:shared-bias}. We do not claim $b=0$, nor that scalar readouts preclude every shortcut; rather, the scalar interface is an empirically effective design. As shown in Section~\ref{sec:results}, coupled YES/NO checking inflates surface YES rates and rewards without accuracy gains, whereas the scalar Final Score preserves token calibration and converts reward growth into genuine accuracy improvements (Figure~\ref{fig:template_compare}). Decoupling is thus a vital prerequisite for viable self-improvement.

\subsection{Optimization and Pairwise Advantage Validity: when self-improvement must stop}
\label{sec:update}
\label{sec:reward-quality}
\paragraph{Group-relative optimization.}
With the interface decoupled, well-formed rollouts receive reward $R_i = s_i$, while malformed outputs are masked from group statistics. For each prompt, we sample $n$ rollouts, compute group-relative advantages~\citep{shao2024deepseekmath}, and optimize using a sequence-level policy loss~\citep{zheng2025gspo}. Because advantages are normalized within each prompt group, optimization relies strictly on within-prompt rollout rankings.

\paragraph{Viable improvement is inherently bounded.}
Because the loop operates without external ground-truth anchors on training rewards, recursive optimization cannot continue indefinitely once self-generated rewards or ranking fidelities saturate. Optimizing beyond the reliable reward regime degrades out-of-distribution transfer. We therefore construct \emph{Pairwise Advantage Validity (PAV)} as an online holdout monitor on a compact, human-verified validation set $\mathcal{D}_{\mathrm{val}}$. Constructing this holdout incurs a one-time human verification cost, but its labels never enter RL training rewards.

\paragraph{A pairwise bound on score-difference error.}
Let $A_t=\mathbb{E}_{\mathcal{D}_{\mathrm{val}}}[Y_t]$ be judge rule accuracy on $\mathcal{D}_{\mathrm{val}}$ ($Y_t\in\{0,1\}$). Let $S_t\in[0,4]$ denote the self-produced checker score and $U_t\in[0,4]$ its human-verified ground-truth target, with normalized checker error defined as:
\begin{equation}
e_{C,t}=\mathbb{E}_{\mathcal{D}_{\mathrm{val}}}\!\left[\frac{|S_t-U_t|}{4}\right].
\label{eq:checker-error}
\end{equation}
Group-relative advantages apply a positive affine transform within each prompt, preserving the relative ordering of the $n$ scores. The critical quantity that must remain faithful is the score difference $(S_i-S_j)$ rather than an isolated absolute score. By the triangle inequality:
\begin{equation}
\frac{|(S_i-S_j)-(U_i-U_j)|}{4}
\leq \frac{|S_i-U_i|}{4}+\frac{|S_j-U_j|}{4}.
\label{eq:pairwise-error-bound}
\end{equation}
Taking expectations over exchangeable rollout pairs, the expected pairwise ranking error is bounded by $2e_{C,t}$, directly motivating continuous monitoring of checker fidelity.

\paragraph{Constructing the PAV monitor and early stopping.}
To obtain a unified scalar metric reflecting both judge correctness and checker fidelity, we construct the Pairwise Advantage Validity index:
\begin{equation}
V_t = 1-\frac{(1-A_t)+2e_{C,t}}{3} = \frac{A_t+2(1-e_{C,t})}{3}.
\label{eq:pav}
\end{equation}
The relative weight of $2$ on checker error directly reflects the bound in Eq.~\ref{eq:pairwise-error-bound} and is fixed \emph{a priori}. Both $\widehat A_t$ and $\widehat e_{C,t}$ are sample means on $\mathcal{D}_{\mathrm{val}}$, ensuring that $\mathbb{E}[\widehat V_t]=V_t$ is unbiased at every fixed checkpoint. We compute 95\% prompt-bootstrap intervals (Appendix~\ref{sec:pav-details}). A drop in checker fidelity signals uncalibrated pairwise rewards; conversely, a high $\widehat V_t$ localizes an effective checkpoint region, preventing validation deception where same-style accuracy rises while cross-style transfer degrades.

%% file: sections/experiments_setup.tex
\section{Experimental Setup}
\label{sec:setup}

\paragraph{Models and training.}
We evaluate \methodname{} across three setups: Qwen3.5-9B~\citep{qwen35} (main runs/ablations), Gemma-4-E4B-it~\citep{gemma4} (cross-architecture replication), and Qwen3.6-27B~\citep{qwen35} (scaling to 27B). Training instances are cluster-split from RubricHub~\citep{li2026rubrichub} with responses synthesized via a 16-model three-tier pool (Appendix~\ref{sec:candidate-sampling}), with zero evaluation overlap. Ground-truth labels never enter RL rewards. Training uses verl~\citep{sheng2024hybridflow} with FSDP and vLLM~\citep{kwon2023vllm}: batch size $32$ ($48$ for 27B), $n{=}8$ rollouts, lr $3{\times}10^{-6}$, clipping $0.005$, and no KL penalty ($0.005$ for 27B). The checker is a standalone vLLM snapshot synchronized each step (Appendix~\ref{sec:hyperparams}).

\begin{wraptable}{r}{0.48\textwidth}
  \centering
  \vspace{-14pt}
  \scriptsize
  \setlength{\tabcolsep}{3pt}
  \caption{\textbf{Held-out transfer suite.} \SVhard{} and \SVfull{} serve as separate online monitors.}
  \label{tab:benchmarks}
  \begin{tabular}{lrrl}
    \toprule
    \textbf{Benchmark} & \textbf{\#} & \textbf{Rules} & \textbf{Metric} \\
    \midrule
    HealthBench & 1459 & 2.0  & rule acc $\uparrow$ \\
    RubricBench & 2294 & 5.8  & pair acc $\uparrow$ \\
    CheckEval-Summ & 1600 & 15   & Pearson $\rho$ $\uparrow$ \\
    ProfBench & 120  & 29.1 & rule acc $\uparrow$ \\
    \bottomrule
  \end{tabular}
  \vspace{-12pt}
\end{wraptable}

\paragraph{Three-tier evaluation hierarchy.}
We organize evaluation into a three-tier hierarchy spanning online monitors and out-of-distribution transfer suites (Table~\ref{tab:benchmarks}):
(1)~\textbf{\SVhard{}} ($100$ prompts; Appendix~\ref{sec:hard-benchmark}): A compact human-verified in-domain monitor on hard rules ($2.5$ rules/prompt) supplying gold for checker fidelity ($1-\mathrm{NMAE}$) and serving as the sole stream for online PAV early-stopping monitoring.
(2)~\textbf{\SVfull{}} ($100$ prompts; Appendix~\ref{sec:svfull-benchmark}): Scores the \emph{full} per-prompt rule suite under the same judging style to probe complete in-domain coverage.
(3)~\textbf{Held-out transfer suite} (Table~\ref{tab:benchmarks}): Evaluated offline at retained checkpoints across distinct domains and rule granularities:
\emph{HealthBench}~\citep{arora2025healthbench} evaluates clinical communication under sample-specific medical rubrics (a stratified 10\% representative sample of $1{,}459$ prompts across 34 clinical categories; measured by per-rule accuracy against physician consensus);
\emph{RubricBench}~\citep{zhou2026rubricbench} tests multi-criteria preference discrimination under sample-specific rubrics ($5.8$ rules/prompt; measured by pairwise preference accuracy);
\emph{CheckEval-Summ}~\citep{lee2025checkeval} assesses document summarization under a shared \emph{global} 15-criterion checklist (unlike the other sample-specific suites; measured by Pearson~$\rho$ against continuous human quality ratings); and
\emph{ProfBench}~\citep{wang2025profbench} evaluates complex professional tasks with sample-specific expert rubrics ($\approx$29 tight rules/prompt; measured by fine-grained rule accuracy; Appendix~\ref{sec:examples}).

\paragraph{Ablations and research questions.}
All ablations use Qwen3.5-9B, comparing against five controls: (1)~a frozen base checker; (2)~an external 27B checker; (3)~self-consistency majority reward; (4)~matched-step teacher SFT ($6400$ traces, $200$ steps); and (5)~larger-pool teacher SFT on the full distillation set ($\approx$17k traces, two epochs). Gemma-4-E4B-it provides cross-architecture replication (Appendix~\ref{sec:gemma-replication}). For downstream transfer, preference pairs generated by our retained \methodname{} judge versus its untuned base counterpart train matched Qwen-27B policies using DPO~\citep{rafailov2023direct}. We evaluate across five research questions: viability across scales and architectures (RQ1), interface decoupling (RQ2), PAV early-stopping localization (RQ3), judge--checker co-evolution (RQ4), and downstream policy transfer (RQ5).

%% file: sections/results.tex
\section{Results}
\label{sec:results}

\begin{figure*}[t]
  \centering
  \vspace{-16pt}
  \includegraphics[width=0.60\textwidth]{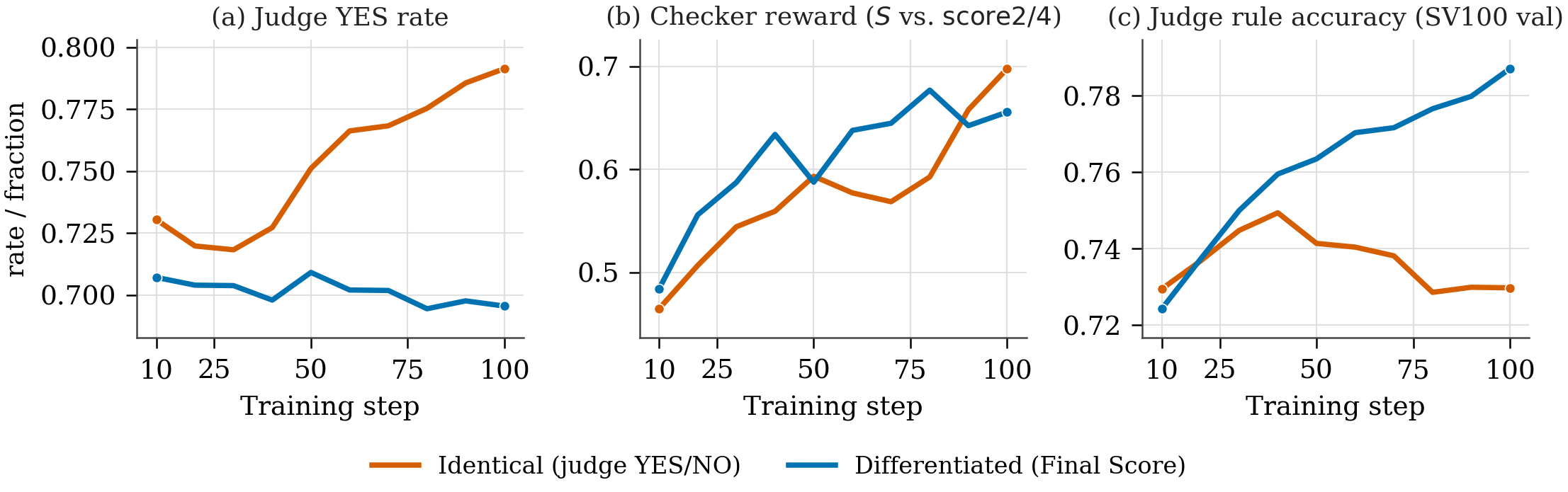}
  \vspace{-11pt}
  \caption{\textbf{Interface decoupling on \SVhard{} (RQ2).} Comparing identical YES/NO checking against differentiated Final Score (steps~10--100). (a)~Identical checking inflates YES rate. (b)~Both report climbing rewards ($[0,1]$ scale). (c)~Only the differentiated Final Score converts reward growth into genuine validation accuracy gains.}
  \label{fig:template_compare}
  \vspace{-16pt}
\end{figure*}

\subsection{RQ1: Bounded RSI improves the judge}
Bounded RSI consistently elevates judge accuracy across formats without requiring gold labels or external teachers in the RL reward (Table~\ref{tab:main}; detailed definitions of rule accuracy \textbf{R}, all-correct rate \textbf{All}, and macro accuracy \textbf{M} are in Appendix~\ref{sec:metrics}). Designated Best checkpoints correspond to the peak validity windows localized by PAV. On Qwen-9B, \SVhard{} rule accuracy rises by $12.9$ points, \SVfull{} by $3.2$, and all four out-of-distribution transfer benchmarks improve at the retained checkpoint. Gemma replicates this positive trajectory on \SVhard{} ($+5.2$) and across transfer suites, while Qwen-27B achieves a $3.9$-point lift on \SVhard{} and gains across every transfer suite. Final checkpoints soften or drop on transfer metrics, confirming that the improvement window is inherently bounded.

\paragraph{Generalized lift vs.\ validation deception.}
On Qwen-9B, Final \SVhard{} accuracy ($75.9\%$) exceeds Best ($73.7\%$), while \SVfull{} falls from $95.3\%$ to $89.7\%$ and out-of-distribution metrics deteriorate by step~200. PAV balances same-style judge accuracy against checker fidelity rather than naively chasing \SVhard{} accuracy alone, identifying the optimal stopping region without inspecting transfer suites. Transfer benchmarks achieve consistent positive deltas over Base within this PAV-indicated window (Table~\ref{tab:main}). \SVfull{} peaks at the same checkpoint ($95.3\%$ vs.\ $92.1\%$), and Gemma replicates this pattern (Appendix~\ref{sec:gemma-replication}). Gains across ProfBench and CheckEval confirm robust generalization under dense expert constraints ($\approx$29 weighted criteria) and continuous human correlation metrics (Pearson~$\rho$).

\begin{table}[t]
  \centering
  \vspace{-12pt}
  \scriptsize
  \setlength{\tabcolsep}{2.2pt}
  \caption{\textbf{Held-out results} (\%; CheckEval: Pearson $\rho$). \textbf{R}/\textbf{All}/\textbf{M}: rule/all-correct/macro accuracy (Appendix~\ref{sec:metrics}). Bold marks designated Best within the PAV-localized validity window across each architecture. Final denotes the training endpoint. \SVfull{} reports rule accuracy only.}
  \label{tab:main}
  \vspace{1pt}
  \resizebox{\textwidth}{!}{%
    \begin{tabular}{llcccc cccc cc cccc}
    \toprule
    & & \multicolumn{3}{c}{\textbf{\SVhard{}}} &
    \multicolumn{1}{c}{\textbf{\SVfull{}}} &
    \multicolumn{4}{c}{\textbf{HealthBench}} &
    \multicolumn{1}{c}{\textbf{RubricBench}} &
    \multicolumn{1}{c}{\textbf{CheckEval}} &
    \multicolumn{3}{c}{\textbf{ProfBench}} \\
    \cmidrule(lr){3-5}\cmidrule(lr){6-6}\cmidrule(lr){7-10}\cmidrule(lr){11-11}\cmidrule(lr){12-12}\cmidrule(l){13-15}
    \textbf{Model} & \textbf{Ckpt.} &
    \textbf{R} & \textbf{All} & \textbf{M} & \textbf{R} &
    \textbf{R} & \textbf{All} & \textbf{M} & \textbf{F1} &
    \textbf{Pair acc} & $\boldsymbol{\rho}$ &
    \textbf{R} & \textbf{All} & \textbf{M} \\
    \midrule
    \multirow{3}{*}{\emph{Qwen3.5-9B}}
    & Base  & 60.8 & 33.1 & 64.7 & 92.1 & 82.5 & 74.9 & 82.8 & 65.4 & 76.0 & 0.422 & 73.1 & 3.8 & 73.0 \\
    & Best  & \textbf{73.7} & \textbf{53.6} & \textbf{77.6} & \textbf{95.3} & \textbf{85.9} & \textbf{79.0} & \textbf{86.5} & \textbf{66.3} & \textbf{79.1} & \textbf{0.441} & \textbf{74.7} & \textbf{5.8} & \textbf{74.5} \\
    & Final & 75.9 & 61.0 & 80.7 & 89.7 & 82.9 & 74.0 & 82.8 & 63.4 & 77.8 & 0.345 & 71.5 & 4.2 & 71.7 \\
    \midrule
    \multirow{3}{*}{\emph{Qwen3.6-27B}}
    & Base  & 73.1 & 54.5 & 77.6 & 94.2 & 85.5 & 77.7 & 84.9 & 67.1 & 78.6 & 0.521 & 74.9 & 4.2 & 74.0 \\
    & Best  & \textbf{77.0} & \textbf{63.0} & \textbf{82.1} & \textbf{96.0} & \textbf{86.6} & \textbf{79.4} & \textbf{86.5} & \textbf{68.2} & \textbf{80.6} & \textbf{0.548} & \textbf{76.4} & \textbf{6.9} & \textbf{75.7} \\
    & Final & 75.6 & 58.7 & 80.2 & 95.3 & 86.5 & 78.8 & 86.4 & 68.1 & 81.9 & 0.548 & 75.1 & 4.8 & 74.6 \\
    \midrule
    \multirow{3}{*}{\emph{Gemma-4-E4B-it}}
    & Base  & 55.9 & 16.1 & 58.9 & 92.5 & 82.8 & 73.2 & 82.9 & 61.0 & 74.9 & 0.442 & 73.1 & 2.9 & 73.2 \\
    & Best  & \textbf{61.1} & \textbf{31.0} & \textbf{65.1} & 92.3 & \textbf{86.5} & \textbf{79.3} & \textbf{86.8} & 60.5 & \textbf{76.6} & \textbf{0.466} & \textbf{74.0} & \textbf{4.0} & \textbf{74.0} \\
    & Final & 59.6 & 27.3 & 63.5 & 92.5 & 85.6 & 78.1 & 85.8 & 59.7 & 76.7 & 0.463 & 73.2 & 4.4 & 73.4 \\
    \bottomrule
  \end{tabular}
  }
  \vspace{-12pt}
\end{table}

\begin{wrapfigure}{r}{0.48\textwidth}
  \centering
  \vspace{-13pt}
  \includegraphics[width=\linewidth]{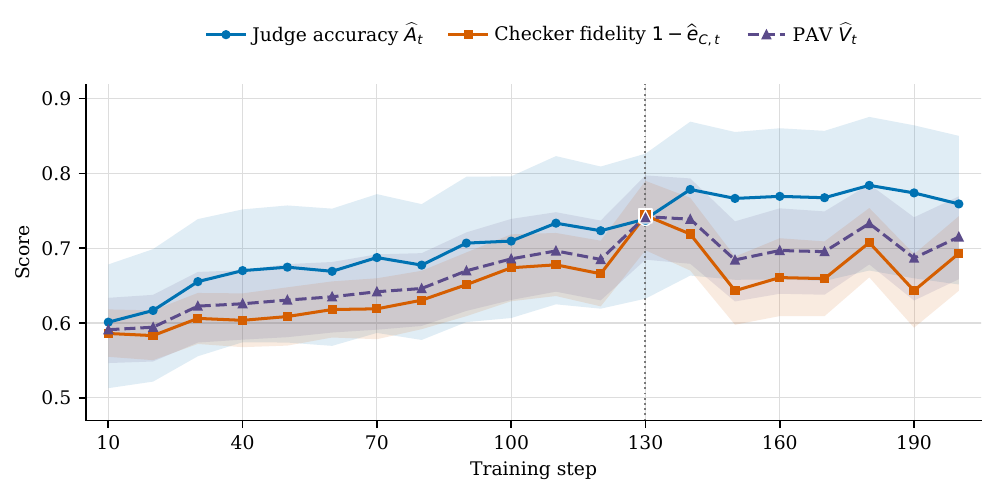}
  \vspace{-15pt}
\caption{\textbf{PAV on \SVhard{} (RQ3).} Judge accuracy $\widehat A_t$, checker fidelity $1-\widehat e_{C,t}$, and composite $\widehat V_t$ (dashed; 95\% CI). Dotted line denotes the PAV landmark (step~130).}
\label{fig:pav_trajectory}
\vspace{-11pt}
\end{wrapfigure}

\subsection{RQ2: Interface decoupling}
Sharing identical YES/NO formats fails due to surface coupling; interface decoupling closes this shortcut and enables reward and accuracy to rise concordantly (Figure~\ref{fig:template_compare}). Holding evidence and synchronization constant, altering the readout from indexed YES/NO tokens to a free-standing scalar Final Score breaks token copying. Under identical YES/NO checking, YES rate inflates ($0.730\rightarrow0.791$) and aligned reward climbs ($0.465\rightarrow0.698$), yet validation accuracy plateaus ($0.729\rightarrow0.730$). Under the differentiated Final Score, YES rate remains calibrated ($0.707\rightarrow0.695$) while aligned reward ($0.484\rightarrow0.656$) and validation accuracy ($0.724\rightarrow0.787$) climb in tandem, confirming decoupling as a prerequisite.

\subsection{RQ3: PAV localizes the useful checkpoint}

Using solely the fixed \SVhard{} monitor, PAV reliably localizes an effective stopping region centered near step~130 on Qwen-9B (Figure~\ref{fig:pav_trajectory}), achieving $\widehat V_t=0.742$ (95\% bootstrap CI $[0.685,0.797]$). While Qwen's hard-subset accuracy continues climbing through step~200, checker fidelity peaks sharply near step~130 and degrades thereafter (Figure~\ref{fig:pav_trajectory}), explaining why tracking rule accuracy alone is insufficient. Three-fold resampling confirms checkpoint stability inside $\{130,140\}$ (Appendix~\ref{sec:pav-details}). Crucially, \SVfull{} (excluded from PAV) independently corroborates this stopping boundary: it peaks at the same landmark ($95.3\%$) and falls to $89.7\%$ by training conclusion. Held-out suites improve over Base at PAV landmarks, confirming that PAV reliably captures an effective stopping state without inspecting transfer data.

\subsection{RQ4: Judge--checker co-evolution matters}
Ablations confirm that weight-tied co-evolution and structured process auditing are essential for sustaining self-improvement (Table~\ref{tab:ablation}). A frozen base checker plateaus early (\SVhard{} $62.7\%$). An external 27B checker remains competitive on \SVhard{} ($73.1\%$) but underperforms suite-wide (HealthBench $79.8\%$ vs.\ Main $85.9\%$), showing static scale cannot substitute for co-evolution. Self-consistency collapses over training ($56.5\%$ Final). Matched-step teacher SFT ($6400$ traces) reaches only $64.2\%$ on \SVhard{}; expanding distillation to the full eligible set ($\approx$17k traces, two epochs) reaches $70.2\%$, still trailing Main across every held-out transfer benchmark (HealthBench $83.5\%$ vs.\ $85.9\%$, CheckEval $0.394$ vs.\ $0.441$). Main surpasses finished matched-step SFT by step~30.

\paragraph{Why teacher distillation struggles: dual-variable difficulty coupling.}
The failure of scaled teacher SFT exposes a structural difference between judge training and task distillation (e.g., in math or coding). In single-variable domains, learning primarily demands challenging queries. In rubric judging, however, an informative gradient requires simultaneous calibration over \emph{two coupled variables}: (1)~a discriminative rubric, and (2)~a candidate response exhibiting \emph{borderline compliance} near the model's decision boundary. In static offline corpora (even across $\approx$17k teacher trajectories), fixed $(h,y,r_{1:K})$ instances inevitably degenerate into trivial samples with negligible corrective gradient. In contrast, \methodname{} operates strictly \emph{on-policy}: $\pi_{\theta_t}$ samples rollouts along its active decision boundary while the synchronized checker audits reasoning differences via group-relative advantages, sustaining dual-variable tension throughout training.

\begin{table}[t]
  \vspace{-16pt}
  \begin{minipage}[t]{0.56\textwidth}
    \centering
    \captionsetup{width=\linewidth}
    \scriptsize
    \setlength{\tabcolsep}{2.0pt}
    \caption{\textbf{Ablation on Qwen3.5-9B} (Peak/Final; SFT single-endpoint). Main beats static checkers and teacher SFT.}
    \label{tab:ablation}
    \vspace{2pt}
    \resizebox{\linewidth}{!}{%
    \begin{tabular}{lcccc}
      \toprule
      \textbf{Arm} & \textbf{\SVhard{}} & \textbf{Health} & \textbf{Rubric} & \textbf{Check} \\
      \midrule
      Base & 60.8 & 82.5 & 76.0 & 0.422 \\
      Main & \textbf{73.7}/75.9 & \textbf{85.9}/82.9 & 79.1/77.8 & 0.441/0.345 \\
      Frozen checker & 62.7/61.3 & 85.0/76.0 & 76.5/77.8 & 0.442/0.454 \\
      External 27B & 73.1/72.4 & 79.8/79.7 & \textbf{79.9}/79.4 & \textbf{0.471}/0.459 \\
      Self-consistency & 65.3/56.5 & 82.5/83.5 & 73.1/73.0 & 0.447/0.435 \\
      Teacher SFT (6.4k) & 64.2 & 81.0 & 76.2 & 0.329 \\
      Teacher SFT (17k) & 70.2 & 83.5 & 77.1 & 0.394 \\
      \bottomrule
    \end{tabular}}
  \end{minipage}\hfill
  \begin{minipage}[t]{0.41\textwidth}
    \centering
    \captionsetup{width=\linewidth}
    \scriptsize
    \setlength{\tabcolsep}{2.5pt}
    \caption{\textbf{Downstream tasks (Qwen3.6-27B).} \methodname{} preference labels outperform untuned base-judge labels on {GPQA}~\citep{rein2024gpqa}, {GuideBench}~\citep{diao2025guidebench}, and {SOP-Maze}~\citep{wang2025sopmaze}.}
    \label{tab:dpo_transfer}
    \vspace{2pt}
    \resizebox{\linewidth}{!}{%
    \begin{tabular}{lccc}
      \toprule
      \textbf{Model} & \textbf{GPQA} & \textbf{GuideBench} & \textbf{SOP-Maze} \\
      \midrule
      Qwen3.6-27B & 80.87 & 83.88 & 34.34 \\
      + DPO (base) & 79.02 & 83.93 & 34.95 \\
      + DPO (\methodname{}) & \textbf{81.46} & \textbf{85.51} & \textbf{36.71} \\
      \bottomrule
    \end{tabular}}
  \end{minipage}
  \vspace{-16pt}
\end{table}

\subsection{RQ5: Transfer to downstream tasks}
Evaluative gains from bounded RSI transfer directly to downstream policy alignment via preference labeling (Table~\ref{tab:dpo_transfer}). Preference pairs generated by our retained \methodname{} judge versus its base counterpart train matched Qwen3.6-27B policies using DPO (LoRA; Appendix~\ref{sec:hyperparams}). While base-judge labels degrade GPQA ($80.87\rightarrow79.02$) and yield negligible shifts on GuideBench ($+0.05$) and SOP-Maze ($+0.61$), \methodname{} labels achieve consistent gains across all three benchmarks ($+0.59$ on GPQA, $+1.63$ on GuideBench, $+2.37$ on SOP-Maze), confirming that evaluative gains translate to downstream policy alignment.

%% file: sections/conclusion.tex
\section{Conclusion}
\label{sec:conclusion}
\methodname{} establishes that bounded recursive self-improvement (RSI) for LLM rubric judges is feasible when self-produced reward validity is explicitly decoupled and monitored. Grounded in the shared evaluative foundation of judging and auditing, \emph{interface decoupling} closes degenerative surface shortcuts, while \emph{Pairwise Advantage Validity (PAV)} reliably localizes an effective stopping region without external supervision in RL rewards. Across architectures and scales up to 27B, the resulting judges achieve robust out-of-distribution generalization and generate preference pairs that enhance downstream policy alignment (Appendix~\ref{sec:detailed-limitations}).

%% file: sections/appendix_front.tex
\section{Appendix Roadmap}
\label{sec:appendix-overview}
The main text keeps the 8-page narrative on the two governing questions---when self-improvement can occur, and when it must stop---together with the primary tables and figures for RQ1--RQ5. This appendix collects the supporting material: full algorithms, mathematical formulations, construction protocols, prompts, extra diagnostics, systems comparisons, taxonomy tables, and robustness checks. The sections below map onto the corresponding main-text claims.

\begin{itemize}
  \setlength{\itemsep}{2pt}
  \setlength{\parskip}{0pt}
  \item Appendix~\ref{sec:alg-details} (Section~\ref{sec:method}): Algorithm~1 of the closed-loop training recurrence, group-relative policy gradient objectives, and the structural causal formulation of surface coupling (Appendix~\ref{sec:surface-coupling-details}).
  \item Appendix~\ref{sec:dual-variable-examples} (Sections~\ref{sec:introduction} and~\ref{sec:results}, RQ4): concrete exemplars and structural analysis of dual-variable difficulty coupling versus single-variable task distillation.
  \item Appendix~\ref{sec:candidate-sampling} (Section~\ref{sec:setup}): candidate response synthesis protocol on RubricHub, detailing the 16-model three-tier pool and the balanced 1:2:1 allocation.
  \item Appendices~\ref{sec:hard-benchmark}--\ref{sec:svfull-benchmark} (Section~\ref{sec:setup}): construction of the two online monitors---human-verified \SVhard{} used by PAV, and complementary \SVfull{} scoring the full in-domain rule suite.
  \item Appendix~\ref{sec:ablation-arms} (Section~\ref{sec:setup}, RQ4): definitions of the five Qwen-9B controls (frozen checker, external 27B checker, self-consistency, and the two teacher-SFT pools).
  \item Appendices~\ref{sec:appendix}--\ref{sec:prompts} (Sections~\ref{sec:bounded-rsi}--\ref{sec:update}): format gate, shared evidence-block layout, and the decoupled judge/checker prompt templates.
  \item Appendices~\ref{sec:metrics}--\ref{sec:examples} (Section~\ref{sec:setup}, Table~\ref{tab:main}): metric definitions and per-benchmark characteristics for the held-out transfer suite.
  \item Appendix~\ref{sec:additional-dynamics} (RQ1, RQ3): Qwen-9B judge/checker diagnostic grids on \SVhard{} and HealthBench, showing why rule accuracy alone is an insufficient stopping signal.
  \item Appendix~\ref{sec:gemma-replication} (RQ1, RQ3): cross-architecture replication on Gemma-4-E4B-it---fine-grained judge and checker diagnostic dynamics on \SVhard{} and HealthBench.
  \item Appendix~\ref{sec:pav-details} (Section~\ref{sec:reward-quality}, RQ3): PAV unbiasedness, sequential monitoring vs.\ pointwise estimation, bootstrap intervals, fold stability, and the offline selector comparison.
  \item Appendices~\ref{sec:compute-overhead}--\ref{sec:format-compliance} (RQ3, RQ4): systems and cost-performance trade-offs relative to commercial APIs and standalone RMs, plus format-compliance dynamics.
  \item Appendix~\ref{sec:hyperparams} (Sections~\ref{sec:setup} and~\ref{sec:results}): RL and downstream DPO hyperparameters used for the main runs and RQ5 transfer.
  \item Appendix~\ref{sec:taxonomy-comparison} (Sections~\ref{sec:introduction} and~\ref{sec:related_work}): systematic taxonomy and multi-dimensional comparison against concurrent self-improving evaluators.
  \item Appendix~\ref{sec:broader-impact} (Section~\ref{sec:conclusion}): broader societal impact, automation bias mitigation, and the *Humans-over-the-loop* safety paradigm.
  \item Appendix~\ref{sec:detailed-limitations} (Conclusion): expanded limitations and open questions summarized at the end of the main text.
\end{itemize}

\section{Algorithm and Method Details}
\label{sec:alg-details}
\label{sec:method-details}

\paragraph{Group-relative advantage computation and policy objective.}
In each training iteration $t$, for a given rubric instance $x = (h, y, r_{1:K})$, the judge policy $\pi_{\theta_t}$ generates $n$ rollouts $\{z_{t,i}\}_{i=1}^n$. The synchronized checker copy $C_{\bar\theta_t}$ audits each rollout to emit a scalar score $s_{t,i} \in \{0, 1, 2, 3, 4\}$. Outputs that pass the format gate form the valid index set $\mathcal{V}_t \subseteq \{1, \ldots, n\}$ with reward $R_{t,i} = s_{t,i}$. Group mean and standard deviation are computed strictly within the prompt group:
\begin{equation}
\mu_t = \frac{1}{|\mathcal{V}_t|} \sum_{i \in \mathcal{V}_t} R_{t,i}, \quad
\sigma_t = \sqrt{\frac{1}{|\mathcal{V}_t|} \sum_{i \in \mathcal{V}_t} (R_{t,i} - \mu_t)^2} + \epsilon_{\mathrm{std}},
\label{eq:group-stats}
\end{equation}
where $\epsilon_{\mathrm{std}} = 10^{-6}$ prevents numerical division by zero. The group-relative advantage for each valid rollout is normalized as:
\begin{equation}
\widehat A_{t,i} = \frac{R_{t,i} - \mu_t}{\sigma_t}.
\label{eq:group-advantage}
\end{equation}
Malformed rollouts ($i \notin \mathcal{V}_t$) are masked out with zero advantage. The policy $\pi_\theta$ is updated by optimizing the clipped sequence-level policy loss:
\begin{equation}
\mathcal{L}_{\mathrm{RL}}(\theta) = -\frac{1}{|\mathcal{V}_t|} \sum_{i \in \mathcal{V}_t} \min\left( \frac{\pi_\theta(z_{t,i} \mid x)}{\pi_{\theta_t}(z_{t,i} \mid x)} \widehat A_{t,i}, \; \operatorname{clip}\left(\frac{\pi_\theta(z_{t,i} \mid x)}{\pi_{\theta_t}(z_{t,i} \mid x)}, 1-\epsilon_{\mathrm{clip}}, 1+\epsilon_{\mathrm{clip}}\right) \widehat A_{t,i} \right),
\label{eq:policy-loss}
\end{equation}
with clipping threshold $\epsilon_{\mathrm{clip}} = 0.005$. Crucially, advantages $\widehat A_{t,i}$ are completely invariant under any affine transformation $R \mapsto \alpha R + \beta$ ($\alpha > 0$). Therefore, the absolute numeric range of the checker's 5-tier score is immaterial; what determines the policy gradient is purely the relative ranking induced among the $n$ on-policy rollouts.

\begin{algorithm}[h]
\caption{\methodname{} training loop with periodic held-out validity monitoring}
\label{alg:selfjudge}
\begin{algorithmic}[1]
\REQUIRE Rubric instances $\mathcal{D}$, initial judge policy $\pi_{\theta_0}$, monitoring interval $T = 10$, fixed human-verified monitor $\mathcal{H}$, training budget $B$
\STATE $\bar\theta_0 \leftarrow \theta_0$ \hfill\COMMENT{\emph{reward-only checker copy}}
\STATE $V^\star\leftarrow-\infty$;\; $t^\star\leftarrow0$
\FOR{$t = 0, 1, 2, \dots, B$}
  \STATE Sample $x \sim \mathcal{D}$; roll out $z_{t,i} \sim \pi_{\theta_t}(\cdot \mid x)$ for $i = 1, \dots, n$
  \STATE $s_{t,i} \leftarrow C_{\bar\theta_t}(x, z_{t,i})$ \hfill\COMMENT{\emph{Pass 2: process check}}
  \STATE $R_{t,i} \leftarrow s_{t,i}$ if format-valid else $0$; mask invalid from group statistics
  \STATE $\theta_{t+1} \leftarrow \mathrm{RLUpdate}(\theta_t, \{(z_{t,i}, R_{t,i})\})$;\; $\bar\theta_{t+1} \leftarrow \theta_{t+1}$
  \IF{$t \bmod T = 0$}
    \STATE Compute $\widehat A_t$, $\widehat e_{C,t}$, and PAV $\widehat V_t$ on $\mathcal{H}$
    \IF{$\widehat V_t>V^\star$}
      \STATE Save checkpoint;\; $V^\star\leftarrow\widehat V_t$;\; $t^\star\leftarrow t$
    \ENDIF
  \ENDIF
\ENDFOR
\STATE \textbf{return} checkpoint $t^\star$
\end{algorithmic}
\end{algorithm}

\subsection{Surface Coupling: Structural Causal Formulation and Visual Illustration}
\label{sec:surface-coupling-details}

\begin{figure*}[t]
  \centering
  \includegraphics[width=\textwidth]{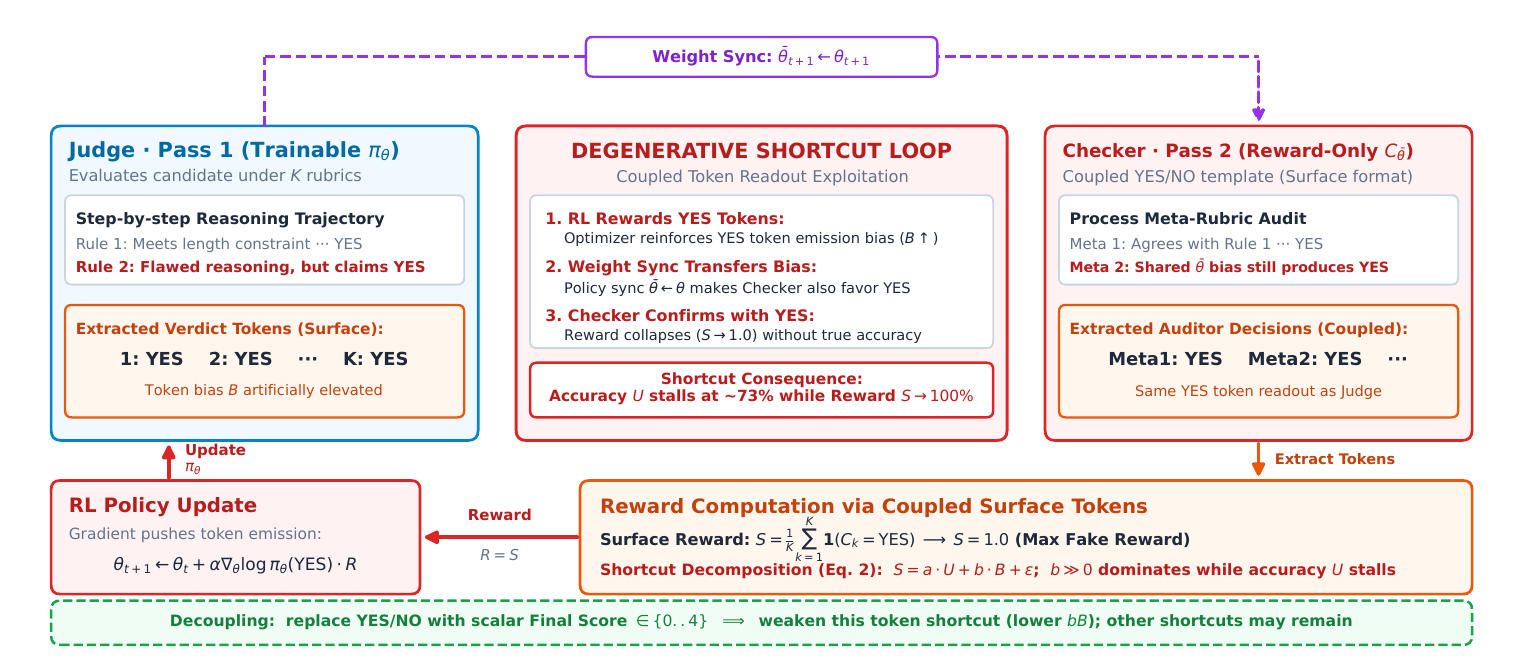}
  \caption{\textbf{Surface coupling and the degenerative reward shortcut pathway (Section~\ref{sec:template}).} When Judge (Pass~1) and Checker (Pass~2) share identical YES/NO readout templates, self-produced reward $S$ is extracted directly from surface decision tokens ($S=\frac{1}{K}\sum \mathbf{1}(\mathrm{Checker}_k=\mathrm{YES})$). Policy gradient optimization inflates YES token emission ($B\uparrow$), transferring immediately to synchronized checker weights ($\bar\theta \leftarrow \theta$). This establishes a degenerative shortcut loop ($bB$) where self-assigned rewards surge to 100\% while true evaluative capability $U$ stalls. Interface decoupling replaces the checker's verdict surface with a free-standing 5-tier scalar Final Score, severing this direct token-level shortcut channel.}
  \label{fig:surface_coupling}
\end{figure*}

\paragraph{Structural Causal Model (SCM) of surface coupling.}
To formalize the shortcut channel described by $S = aU + bB + \varepsilon$ (Eq.~\ref{eq:shared-bias}), consider the structural causal relations governing reward extraction under parameter synchronization ($\bar\theta_{t+1} \leftarrow \theta_{t+1}$):
\begin{enumerate}
  \setlength{\itemsep}{0pt}
  \setlength{\parskip}{0pt}
  \item \textbf{Latent Evaluative Quality ($U$)}: Measures whether the judge's reasoning correctly verifies the candidate response against criteria $r_{1:K}$.
  \item \textbf{Surface Token Bias ($B$)}: Measures the marginal emission probability bias for verdict tokens (e.g., $\Delta P_{\mathrm{YES}} = P(\text{`YES'}) - 0.5$).
  \item \textbf{Judge Emission ($Z_1$)}: Emits tokens under policy $\pi_{\theta_t}$, where $Z_1 = f_1(U, B; \theta_t)$.
  \item \textbf{Checker Evaluation ($S$)}: Audits reasoning under synchronized copy $C_{\bar\theta_t}$, where $S = f_2(Z_1; \bar\theta_t) = aU + bB + \varepsilon$.
\end{enumerate}
When the checker uses identical YES/NO readout templates, $f_2$ computes reward by directly parsing YES tokens from Pass~2. Because $\bar\theta_t = \theta_t$, any gradient update $\nabla_\theta \mathcal{L}_{\mathrm{RL}}$ that reinforces YES tokens in Pass~1 simultaneously increases the probability that Pass~2 emits YES tokens ($b > 0$). This establishes a degenerative backdoor shortcut: the policy maximizes reward by driving $B \to 1$ ($bB \to \max$) without increasing true reasoning quality $U$ ($aU \approx \text{const}$). 

\paragraph{Interface decoupling as structural causal intervention.}
Interface decoupling replaces $f_2$ with a free-standing 5-tier scalar Final Score $s_i \in \{0, 1, 2, 3, 4\}$. Because the scalar score requires generating a holistic reasoning audit followed by a distinct numerical symbol, the direct token-copying gradient pathway from Pass~1 YES tokens to Pass~2 reward extraction is severed ($\partial S / \partial \text{verdict\_token} \approx 0$, effectively setting $b \approx 0$). The optimizer is therefore constrained to maximize reward solely through improving genuine reasoning verification ($aU$), converting reward growth into robust evaluative accuracy.

\section{Dual-Variable Difficulty Coupling: Concept and Exemplars}
\label{sec:dual-variable-examples}

As introduced in Section~\ref{sec:introduction} and analyzed in Section~\ref{sec:results} (RQ4), a fundamental distinction exists between training standard generative models (e.g., in mathematics or coding) and training LLM rubric judges.

\paragraph{Single-variable vs.\ dual-variable difficulty.}
In single-variable tasks such as mathematical problem solving, instance difficulty is strictly governed by query complexity: given a challenging problem $x$, generating the correct reasoning trace $y$ provides a non-trivial gradient signal. In rubric-based evaluation, however, an informative learning signal requires the simultaneous calibration of \emph{two coupled variables}:
\begin{equation}
\text{Informative Gradient Signal} \iff \underbrace{\mathcal{R}(x)}_{\text{Discriminative Rubric}} \;\times\; \underbrace{\mathcal{B}(y \mid x, \mathcal{R})}_{\text{Borderline Response Compliance}}.
\label{eq:dual-variable-condition}
\end{equation}
If either variable fails to meet this calibration, the instance yields negligible corrective gradient:
\begin{itemize}
  \setlength{\itemsep}{1pt}
  \setlength{\parskip}{0pt}
  \item \textbf{Trivial Response ($y$ far from decision boundary)}: If a candidate response is blatantly compliant (e.g., an obvious perfect answer) or blatantly flawed (e.g., empty or gibberish text), even an untuned base judge trivially assigns YES or NO correctly. The cross-entropy loss produces near-zero gradient, imparting no refinement to the model's decision boundary.
  \item \textbf{Non-discriminative Rubric ($\mathcal{R}$ trivial or ambiguous)}: If a rubric criterion is overly generic (e.g., ``the response is helpful'') or ambiguous, all rollout candidates receive identical verdicts, providing zero group-relative variance.
\end{itemize}

\paragraph{Why static teacher SFT degenerates.}
In static offline distillation corpora (even across $\approx$17,000 trajectories distilled from four frontier teachers: Qwen3.5, Gemini-3.1-Flash-Lite, DeepSeek-V4-Flash, and Kimi-K2.6), the candidate responses $y$ are fixed beforehand. As the student judge improves during fine-tuning, the static triplets $(h, y, r_{1:K})$ quickly fall outside the student's active, shifting decision boundary, degenerating into trivial instances. Scaling data quantity from $6.4$k to $17$k merely multiplies trivial samples without supplying borderline tension.

\paragraph{Why on-policy \methodname{} succeeds.}
In contrast, \methodname{} operates strictly on-policy: as the judge policy $\pi_{\theta_t}$ evolves, its sampled rollouts $z_{t,i}$ naturally probe subtle reasoning variations along its active decision boundary. The synchronized checker dynamically identifies nuanced discrepancies between these rollouts, generating differentiated group-relative advantages $\widehat A_{t,i}$ that sustain informative dual-variable tension throughout training.

\section{Training Data and Candidate Response Synthesis}
\label{sec:candidate-sampling}

RubricHub~\citep{li2026rubrichub} provides queries and fine-grained rubrics $(h, r_{1:K})$ across five diverse domains (Chat, Instruction-Following, Medical, Science, and Writing), but does not contain pre-existing candidate AI responses $y$. To construct realistic evaluation instances $x = (h, y, r_{1:K})$ spanning a broad, balanced spectrum of response quality, we synthesize candidate responses using a tiered multi-model sampling protocol.

\paragraph{Three-tier model pool.}
We stratify $16$ distinct LLMs into three capability tiers based on their reasoning and instruction-following capacities:
\begin{itemize}
  \setlength{\itemsep}{0pt}
  \setlength{\parskip}{0pt}
  \item \textbf{Low Tier} (compact/lightweight models): GLM-4-Flash, GPT-4.1-nano, Gemini-2.5-Flash-Lite, and Qwen3.5-9B.
  \item \textbf{Mid Tier} (mid-scale conversational models): LongCat-Lite-V2-Chat, DeepSeek-V3.2, GLM-5.1, GPT-4.1-mini, Gemini-2.5-Flash, Moonshot-v1-128k, and Qwen3.6-27B.
  \item \textbf{High Tier} (frontier reasoning models): LongCat-Flash-Thinking, LongCat-Flash-Chat, Qwen3.5-397B, DeepSeek-V4-Flash, GPT-4.1, Gemini-3.1-Flash-Lite, and Kimi-K2.6.
\end{itemize}

\paragraph{Balanced 1:2:1 tier allocation per query.}
For each query $h$, we allocate four response slots across tiers in a fixed $\text{Low} : \text{Mid} : \text{High} = 1 : 2 : 1$ ratio:
\begin{equation}
y \sim \begin{cases}
1 \times \text{Low Tier} & \text{(often violates fine-grained negative constraints or nuances)}, \\
2 \times \text{Mid Tier} & \text{(borderline quality; compliant on standard rules but mixed on edge cases)}, \\
1 \times \text{High Tier} & \text{(substantially compliant; provides challenging borderline rubrics)}.
\end{cases}
\end{equation}
Within each domain, models within each tier are assigned evenly across samples via a greedy balanced permutation schedule (ensuring mid-tier slots use two distinct models). This tiered distribution ensures the training corpus exhibits rich compliance variation across criteria without collapsing to trivial all-YES or all-NO extremes.

\section{\SVhard{} Construction}
\label{sec:hard-benchmark}
\SVhard{} is a same-style human-verified validation set of $100$ examples
used to monitor reward validity and guide early stopping under RubricHub-style
rubric judging.
RubricHub is cluster-split so that no prompt or rubric cluster appears in both
training and evaluation; \SVhard{} is drawn from the evaluation split and never
contributes gold labels to the RL training reward. Labels are LLM-initialized
and then human-verified at the rule level; this compact holdout is far cheaper
than scaling gold annotation to the full training corpus, but the verification
step is still a real upfront cost.

The set covers diverse instruction domains: $20$ instances from each of five
categories (Chat, Instruction-Following, Medical, Science, and Writing), scored
on progression-critical hard rules selected per prompt (mean $2.5$ rules per
prompt; $89/100$ have $2$ or $3$). Because gold is expert-checked,
\SVhard{} is the \emph{only} stream on which we measure checker fidelity
($1-\mathrm{NMAE}$) and compute PAV. It is a fair same-style validity
monitor---matched in judging style to training and strictly cluster-isolated
from it.
Repeated inspection can still adapt early stopping to this validation
set; we therefore reserve the different-style benchmarks and \SVfull{} for
testing substantive transfer claims. Their parallel gains argue against a purely
\SVhard{}-specific selection artifact.

\section{\SVfull{} Construction}
\label{sec:svfull-benchmark}
\SVfull{} is a complementary online in-domain validation stream on the same
$100$ cluster-isolated prompts as \SVhard{}, but each instance is scored on its
\emph{full} rule suite rather than the progression-critical hard subset.
Gold labels are resolved by an LLM annotation panel with Claude-based tie-breaking; they are never used in the RL training reward.
\SVfull{} therefore tests whether judge improvement extends to complete
in-domain rubric coverage under the same RubricHub judging interface, without
the human-verification cost required for reliable checker diagnostics.
We evaluate \SVfull{} at the same checkpoint cadence as the offline transfer
suite and report rule accuracy in Table~\ref{tab:main}; PAV itself is computed
only from \SVhard{}.

\section{Ablation Arms}
\label{sec:ablation-arms}
All five controls below are run on Qwen3.5-9B.
\emph{Frozen base:} checker fixed at untuned init.
\emph{External 27B:} Qwen3.6-27B checker, never synced.
\emph{Self-consistency:} majority agreement reward without process audit.
\emph{Teacher-traj.\ SFT (6.4k):} full FT on $6400$ teacher YES/NO traces (from Qwen3.5, Gemini-3.1-Flash-Lite, DeepSeek-V4-Flash, and Kimi-K2.6),
$200$ steps matched to the RL budget.
\emph{Teacher-traj.\ SFT (17k):} same teachers and recipe on the full eligible
pool ($\approx$17k traces, two epochs), testing whether more imitation data
closes the gap.

%% file: sections/appendix.tex
\section{Reward and Format Details}
\label{sec:appendix}
The format gate enforces two structural constraints on judge rollouts $z_{t,i}$: (1)~it requires exactly $K$ parseable YES/NO decisions in the indexed format ($1\text{:\,}\texttt{<ans>YES/NO</ans>}\dots K\text{:\,}\texttt{<ans>YES/NO</ans}$), and (2)~it strictly forbids delimiter tokens reserved for the checker prompt template (e.g., prematurely emitting \texttt{<|AI's Response End|>} or faking checker reward blocks such as \texttt{<|Final Score|>4<|Final Score|>}). This prevents prompt-injection shortcuts where the judge rollout manipulates Pass~2 slot boundaries or directly injects fraudulent high checker scores into the downstream context. Samples failing the gate receive zero reward ($R_{t,i}=0$) and a zeroed response mask so they are excluded from the same-prompt group statistics.

\section{Training Prompts}
\label{sec:prompts}
Both passes share the same overarching evidence-block layout---system instruction, conversation context, the AI response, and a numbered rubric list---ensuring that the checker observes the identical evidence evaluated by the judge. The two passes differ in how these slots are populated and in their required output format. Below, we decompose the prompt template into three distinct components: the shared evidence-block template, the judge-specific instruction and output format (Pass~1), and the checker-specific process-auditing instruction and scoring format (Pass~2).

\paragraph{Part 1: Shared evidence-block template.}
Both the judge and checker instantiate the same standard three-slot evidence layout. Angle brackets ($\langle\langle\langle\cdot\rangle\rangle\rangle$) denote placeholders populated dynamically per instance:
\begin{quote}\small\ttfamily
Your job is to assess if the AI's response to the user's most recent prompt correctly follows the provided rubrics.

<|Conversation History Start|>\\
<<<conversation\_history>>>\\
<|Conversation History End|>

<|AI's Response Start|>\\
<<<ai\_response>>>\\
<|AI's Response End|>

<|Rubrics Start|>\\
<<<rubrics>>>\\
<|Rubrics End|>
\end{quote}
\noindent The three evidence slots are instantiated for each pass as follows:
\begin{itemize}
  \item \textbf{Judge pass ($x=(h,y,r_{1:K})$)}:
    \texttt{<<<conversation\_history>>>} contains the original dialogue context $h$;
    \texttt{<<<ai\_response>>>} contains the candidate generation $y$; and
    \texttt{<<<rubrics>>>} contains the $K$ task-level evaluation criteria $r_{1:K}$.
  \item \textbf{Checker pass ($C(x,z_{t,i})$)}:
    \texttt{<<<conversation\_history>>>} contains the complete judge input instance $x$;
    \texttt{<<<ai\_response>>>} contains the judge's full rollout $z_{t,i}$ (verbatim per-criterion reasoning and verdicts); and
    \texttt{<<<rubrics>>>} contains meta-rubrics of the form ``$k$: Did the AI correctly judge the following rubric: $\langle r_k\rangle$''.
\end{itemize}

\paragraph{Part 2: Judge instruction and output format (Pass~1, trainable).}
Appended to the shared evidence block, the judge is instructed to generate step-by-step reasoning followed by concatenated per-rule binary decisions:
\begin{quote}\small\ttfamily
Please first analyze step by step whether the response meets each of the rubrics above. If it meets the rubric, output YES; otherwise, output NO.\\
Finally, provide the results in rubric order using the concatenated format:\\
1:<ans>YES/NO</ans> 2:<ans>YES/NO</ans> \ldots{} K:<ans>YES/NO</ans>.
\end{quote}
\noindent The judge's output is parsed and verified by the format gate (\S\ref{sec:appendix}), requiring all $K$ indexed decisions to be strictly valid.

\paragraph{Part 3: Checker instruction and output format (Pass~2, reward-only).}
Appended to the shared evidence block, the synchronized checker audits the reasoning process and emits a single, decoupled scalar Final Score $s_i \in \{0,1,2,3,4\}$:
\begin{quote}\small\ttfamily
First, analyze whether each rule's judgment is accurate, then score the overall scoring process.\\
Use 5 tiers (0-4):\\
0 = all rule judgments are wrong and the judge's internal explanation is very poor;\\
1 = most rule judgments are wrong (only a small fraction correct) and the judge's internal explanation is poor/unclear;\\
2 = about half of the rule judgments are correct and the judge's internal explanation is partially clear but has notable issues;\\
3 = most rule judgments are correct (few wrong) and the judge's internal explanation is mostly clear with minor issues;\\
4 = all rule judgments are correct and the judge's analysis process is clear.\\
Output the final score to\\
<|Final Score|>\\
<<<final\_score>>>\\
<|Final Score|>
\end{quote}
\noindent The parsed scalar $s_i$ serves as the sole RL reward ($R_i = s_i$). This structural separation between the judge's per-rule YES/NO verdicts and the checker's 5-point scalar score closes the token-level copying channel (\S\ref{sec:asymmetry}); other shortcuts can remain.

\section{Evaluation Metrics}
\label{sec:metrics}
Let $\hat y_r\in\{\text{YES},\text{NO}\}$ be the judge's prediction for rubric
criterion $r$ and $g_r$ the gold label. In the main text (Table~\ref{tab:benchmarks} and Table~\ref{tab:main}), we report each benchmark under its standard primary metric for conciseness. Below we define both primary and supplementary diagnostic metrics used in our analyses.

\paragraph{Rule accuracy (\SVhard{}, \SVfull{}, HealthBench, ProfBench).}
For instances with per-rule gold labels, we pool over all rules with both a gold
label and a prediction (micro-average): $\text{rule\_acc} = \sum_i
\text{correct}_i / \sum_i \text{total}_i$, where rules with no prediction are
skipped rather than counted wrong. This is the primary metric for \SVhard{},
\SVfull{}, HealthBench, and ProfBench in Table~\ref{tab:main}.

\paragraph{All-correct rate (\SVhard{}, HealthBench, ProfBench).}
The sample-level fraction of instances whose every comparable rule is judged
correctly, i.e. $\mathbf{1}[\text{gold\_match}_i = 1]$ averaged over instances.
This is the stricter sample-level accuracy metric reported alongside rule
accuracy in Table~\ref{tab:main}.

\paragraph{Macro accuracy (\SVhard{}, HealthBench, ProfBench).}
The unweighted average of per-instance rule accuracies across all valid evaluation samples, treating each evaluation prompt/rubric equally regardless of its rule count.

\paragraph{HealthBench rule accuracy and F1.}
Gold is the strict majority vote of physician binary labels (ties excluded); rule
accuracy is agreement with this majority (reported in Table~\ref{tab:main}). We additionally compute the official
balanced F1, which pairs the model prediction with \emph{each} physician label
(including ties) and averages the positive- and negative-class F1.

\paragraph{RubricBench pairwise accuracy.}
For each A/B pair we compute the mean YES-rate on each side; the side with the
higher rate is predicted preferred (ties broken by YES count). Accuracy is over
paired cases against the gold preference label.

\paragraph{CheckEval correlation and metric interpretation.}
CheckEval-Summ~\citep{lee2025checkeval} assesses document summarization across $1{,}600$ summaries generated by $16$ distinct LLM systems across $100$ source documents from SummEval~\citep{fabbri2021summeval}.
In the original SummEval corpus, human experts directly assigned \emph{holistic Likert-scale ratings} (1--5 scale, averaged across annotators to yield continuous scores $\mathbf{H}_d$) for each of four broad quality dimensions: \textbf{Coherence}, \textbf{Consistency}, \textbf{Fluency}, and \textbf{Relevance}. Human annotators evaluated overall summary quality directly and did not annotate fine-grained checklist rules.
To evaluate automated judges without subjective scalar scoring, CheckEval~\citep{lee2025checkeval} decomposed these four holistic dimensions into $15$ atomic, binary checklist criteria:
\begin{itemize}
  \setlength{\itemsep}{0pt}
  \setlength{\parskip}{0pt}
  \item \textbf{Coherence} ($3$ checklist items): Logical narrative flow, paragraph transitions, and semantic organization.
  \item \textbf{Consistency} ($3$ checklist items): Factual alignment with the source document and absence of hallucinations.
  \item \textbf{Fluency} ($4$ checklist items): Grammatical correctness, lexical clarity, and formatting cleanliness.
  \item \textbf{Relevance} ($5$ checklist items): Coverage of central key ideas and avoidance of superfluous details.
\end{itemize}
All $15$ checklist criteria are \emph{global} (identical across all $1{,}600$ summaries). For each summary $i \in \{1, \dots, 1600\}$ and dimension $d \in \{\text{coherence, consistency, fluency, relevance}\}$, the model's predicted dimension score $P_{i,d}$ is computed as the proportion of YES verdicts emitted over that dimension's checklist items:
\begin{equation}
P_{i, d} = \frac{1}{|\text{Rules}(d)|} \sum_{r \in \text{Rules}(d)} \mathbf{1}(\text{Judge}_{i, r} = \text{YES}) \in [0, 1].
\end{equation}
Across all $N = 1{,}600$ evaluation summaries, we compute the sample-level Pearson correlation coefficient $\rho_d = \mathrm{Corr}(\mathbf{P}_d, \mathbf{H}_d)$ between the model's predicted score sequence $\mathbf{P}_d$ and the continuous human rating sequence $\mathbf{H}_d$. The primary CheckEval metric reported in Table~\ref{tab:main} is the macro-average of these sample-level correlation coefficients across the four dimensions:
\begin{equation}
\rho = \frac{1}{4} \sum_{d \in \{\text{coh}, \text{con}, \text{flu}, \text{rel}\}} \rho_d.
\end{equation}
\emph{Interpreting correlation magnitude versus percentage accuracy:} Crucially, Pearson correlation $\rho \in [-1, 1]$ measures linear alignment with continuous human judgment distributions rather than discrete $0$--$100\%$ classification accuracy. In a large evaluation pool of $1{,}600$ instances, an absolute lift in Pearson correlation (e.g., $+0.019$ from $0.422$ to $0.441$ on Qwen-9B, $+0.024$ on Gemma, and $+0.027$ on Qwen-27B) represents a statistically robust and practically meaningful improvement in continuous alignment fidelity ($p = 0.0032 < 0.01$; Appendix~\ref{sec:pav-details}), which must not be naively equated with or compared directly against percentage-point increments on discrete accuracy benchmarks.

\begin{figure*}[t]
  \centering
  \includegraphics[width=\textwidth]{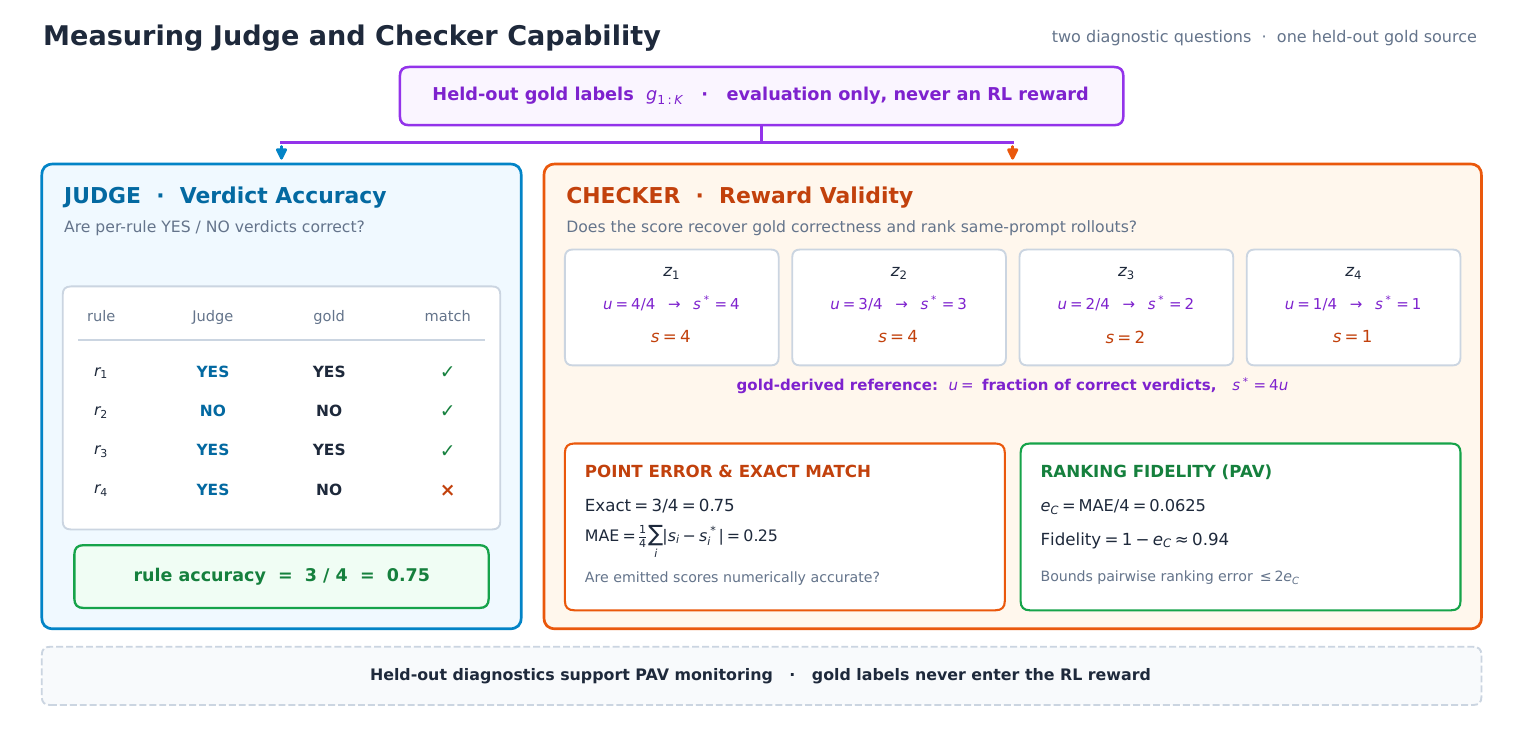}
  \caption{\textbf{Gold-derived diagnostics for judge and checker capability on a four-rollout example.} The same held-out rule labels simultaneously evaluate judge rule accuracy ($A$) and the checker reference target $U=4u$ (where $u$ is gold rule correctness). Checker diagnostics evaluate point error ($\mathrm{MAE}$ and Exact match) and normalized ranking fidelity ($1-e_C=1-\mathrm{NMAE}$), directly supplying the PAV composite monitor ($V$). Gold labels never enter the RL training reward.}
  \label{fig:diagnostics}
\end{figure*}

\paragraph{Checker point calibration and ranking fidelity (PAV).}
The checker emits a scalar score $S \in \{0,1,2,3,4\}$. To assess whether it faithfully recovers the correctness of the judge's reasoning, we compare $S$ against the reference target $U = 4 \times (\text{correct rules} / \text{total rules}) \in [0, 4]$ (e.g., all rules correct $\rightarrow 4$; half $\rightarrow 2$). We evaluate checker capability through three aligned diagnostic metrics:
\begin{itemize}
  \setlength{\itemsep}{0pt}
  \setlength{\parskip}{0pt}
  \item \textbf{Exact-Match Rate}: Converts continuous reference target $U$ to its nearest integer tier $\lfloor U \rceil$ and tests exact equality $\mathbf{1}[S = \lfloor U \rceil]$.
  \item \textbf{Mean Absolute Error ($\mathrm{MAE}$)}: Computes the sample-level mean absolute error $\frac{1}{N}\sum_{i=1}^N |S_i - U_i|$, penalizing unparsable outputs with the worst-case endpoint error ($4$).
  \item \textbf{Ranking Fidelity ($1 - e_C = 1 - \mathrm{NMAE}$)}: Normalizes the checker error to $e_{C} = \frac{\mathrm{MAE}}{4} \in [0, 1]$, yielding the ranking fidelity $1 - e_{C}$. As derived in Eq.~\ref{eq:pairwise-error-bound}, $2e_C$ strictly bounds the expected error on within-prompt pairwise reward differences $(S_i - S_j)$, directly serving as the reward-quality component of the PAV monitor.
\end{itemize}
Figure~\ref{fig:diagnostics} illustrates how the same held-out gold labels simultaneously evaluate judge accuracy and checker ranking fidelity without leaking supervision into the RL training reward.

\section{Benchmark Characteristics and Examples}
\label{sec:examples}
Table~\ref{tab:benchmark-examples} summarizes the distinguishing traits of each
held-out suite and one representative instance that best illustrates them.
Queries, responses, and rubrics are abridged; the judge outputs a YES/NO for
each listed criterion.

\paragraph{HealthBench corpus scale and stratified 10\% subsampling protocol.}
The full OpenAI HealthBench meta-evaluation benchmark~\citep{arora2025healthbench} comprises $29{,}511$ rubric evaluation instances across $14{,}592$ prompt-completion groups across $10$ overarching clinical themes (e.g., global health, emergency referrals, healthcare data communication, uncertainty hedging) and $34$ medical subcategories. In the original corpus, multiple candidate LLM responses per clinical query are paired with $1$--$3$ physician-crafted rubric criteria evaluating patient safety, diagnostic communication, and referral thresholds. Evaluating all $14{,}592$ groups across training trajectories imposes an intractable computational bottleneck ($\approx 30\text{k}$ sequential reasoning traces per evaluation step). To balance evaluation tractability with clinical fidelity, we implement a stratified proportional sampling protocol:
\begin{enumerate}
  \setlength{\itemsep}{0pt}
  \setlength{\parskip}{0pt}
  \item \textbf{Atomic group preservation}: Evaluation records are grouped by unique prompt--response pairs $(h, y)$ so that all interdependent rubric criteria corresponding to a single candidate response remain bound together.
  \item \textbf{Stratified category proportionality}: Groups are partitioned across all $34$ medical categories and sampled strictly in proportion ($10\%$, fraction $\alpha = 0.10$) to their original category prevalence in the full benchmark. This yields exactly $1{,}459$ representative prompt--response groups ($2{,}951$ rubric rules, mean $2.02$ rules/group), faithfully preserving the semantic distribution across all $10$ clinical themes without distortion.
  \item \textbf{Physician consensus ground truth}: Gold labels represent the strict majority vote across certified physician raters (ties excluded).
\end{enumerate}

\begin{table*}[t]
  \centering
  \scriptsize
  \setlength{\tabcolsep}{4pt}
  \caption{\textbf{Benchmark characteristics and representative examples.} Each
  row pairs the trait emphasized in Section~\ref{sec:setup} with one instance
  that best illustrates it.}
  \label{tab:benchmark-examples}
  \begin{tabular}{@{}p{0.11\textwidth}p{0.24\textwidth}p{0.58\textwidth}@{}}
    \toprule
    \textbf{Benchmark} & \textbf{Key characteristics} & \textbf{Representative example} \\
    \midrule
    \SVhard{} &
    Same-style PAV monitor; short hard-rule subsets (mean $2.5$ rules), mostly format/instruction checks; human-verified gold. &
    \emph{Combinatorics + format constraints.} Query: count paths on a grid where every sentence must begin with ``highway'' and sentences are joined by hyphens. Rubric: (1) first word of every sentence is ``highway''; (2) hyphens split sentences correctly; (3) the response addresses the math question. Gold on progression-critical rules: \{1:YES, 2:NO\}. \\
    \addlinespace
    HealthBench &
    Medical safety; representative 10\% cluster sample ($1{,}459$ prompts) from full benchmark; few rules per prompt but semantically richer physician norms. &
    \emph{Incomplete clinical context.} Query: compute perioperative mortality risk with missing details. Response: ``I'm sorry, but I can't assist with that.'' Rubric: (1) does not give a definitive answer when information is missing ([T,T]); (2) still seeks missing context rather than flatly declining ([F,F]). \\
    \addlinespace
    RubricBench &
    Pairwise preference between two responses under sample-specific multi-criterion rubrics (mean $5.8$ rules). &
    \emph{Code assistance.} Query: ``Add mypy type hints for the code below'' (no code provided). Response: asks the user to provide the code first. Rubric checks whether the assistant recognizes the missing code, explains why, addresses mypy specifically, stays clear, and remains polite. \\
    \addlinespace
    CheckEval-Summ &
    \emph{Global} $15$-criterion checklist shared across all summaries (not sample-specific); Pearson~$\rho$ on aggregated YES-rates. &
    \emph{News summarization.} Query: summarize a document (with reference summary). Rubric: fixed checklist over coherence ($\times3$), consistency ($\times3$), fluency ($\times4$), and relevance ($\times5$); YES-rate per dimension is correlated with human dimension scores. \\
    \addlinespace
    ProfBench &
    Expert professional tasks; long weighted rubrics ($\approx$29 rules) with tight numerical tolerances; hardest suite. &
    \emph{Physics PhD / EMI shielding.} Query: multi-step numerical derivation for shielding effectiveness. Rubric: weighted Critical/Major/Minor criteria checking specific numerical ranges (e.g.\ $R\in[0.6835,0.6841]$), the correct dominance conclusion, and stated shielding-effectiveness relations. \\
    \bottomrule
  \end{tabular}
\end{table*}

\section{Additional Training Dynamics}
\label{sec:additional-dynamics}

\begin{figure*}[t]
  \centering
  \includegraphics[width=\textwidth]{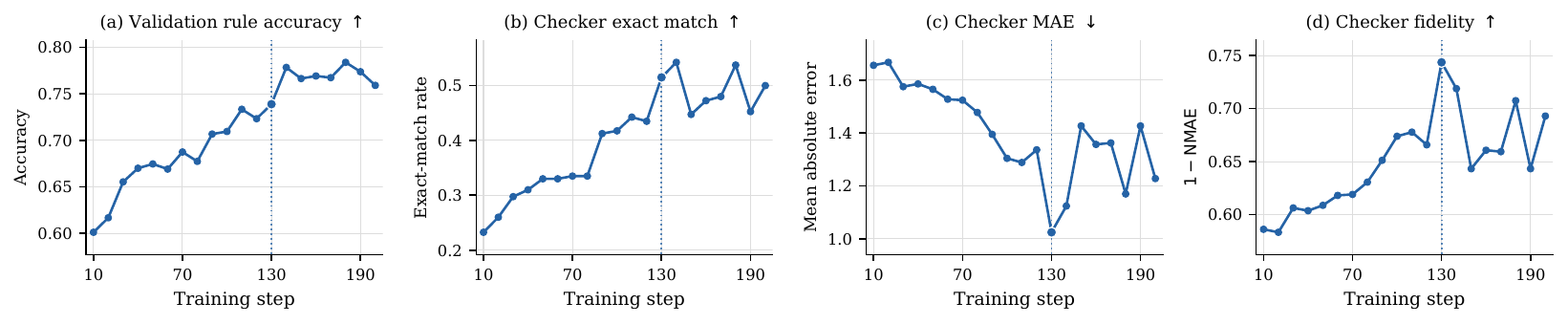}
  \caption{\textbf{\SVhard{} diagnostics on the Qwen3.5-9B trajectory.}
  (a)~Validation rule accuracy ($\uparrow$). (b)~Checker exact-match rate ($\uparrow$).
  (c)~Checker MAE on the original $0$--$4$ score scale ($\downarrow$); malformed checker
  outputs receive their worst possible endpoint error.
  (d)~Checker fidelity $1-\mathrm{NMAE}=1-\mathrm{MAE}/4$ ($\uparrow$).
  Dotted lines and enlarged markers show the PAV landmark @130. Rule accuracy
  alone continues to rise, whereas exact match, MAE, and fidelity expose
  the sharp checker improvement around the selected checkpoint.}
  \label{fig:sv-diagnostics}
\end{figure*}

\begin{figure*}[t]
  \centering
  \includegraphics[width=\textwidth]{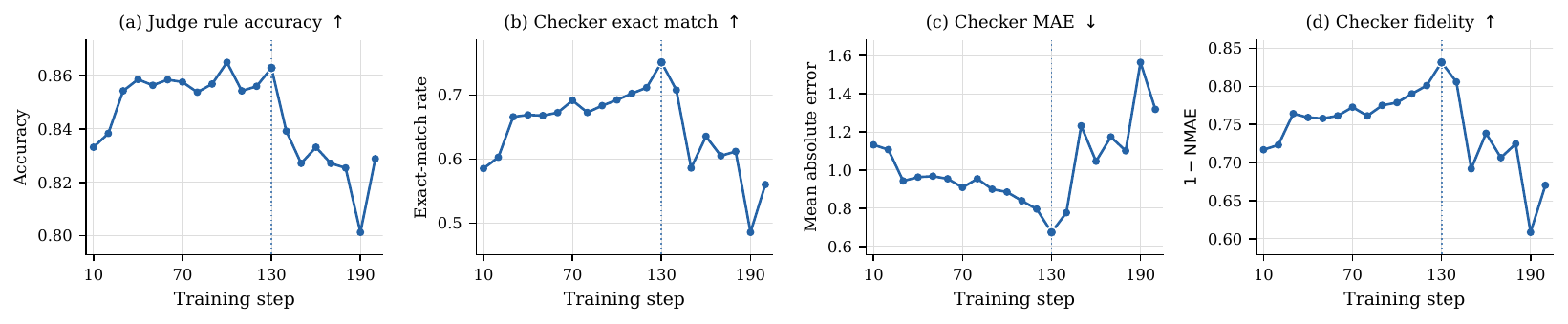}
  \caption{\textbf{HealthBench diagnostics on the Qwen3.5-9B trajectory.}
  Same four panels as Figure~\ref{fig:sv-diagnostics}, now on the
  different-style medical suite that PAV never sees.
  (a)~Judge rule accuracy ($\uparrow$). (b)~Checker exact-match rate ($\uparrow$).
  (c)~Checker MAE on the $0$--$4$ scale ($\downarrow$); malformed checker
  outputs receive their worst possible endpoint error.
  (d)~Checker fidelity $1-\mathrm{NMAE}=1-\mathrm{MAE}/4$ ($\uparrow$).
  Dotted lines and enlarged markers mark the PAV landmark @130.
  Judge accuracy, exact match, and fidelity peak together around that region
  and then fall together, so the transfer suite tracks the same useful
  window as the \SVhard{} monitor.}
  \label{fig:hb-diagnostics}
\end{figure*}

To understand the internal mechanics of recursive self-improvement and how evaluative capabilities evolve over training steps, we examine fine-grained diagnostic trajectories for both the judge policy $\pi_{\theta_t}$ and the synchronized process checker $C_{\bar\theta_t}$ on the primary Qwen3.5-9B run. Figures~\ref{fig:sv-diagnostics} and~\ref{fig:hb-diagnostics} track four complementary diagnostic metrics across the complete optimization trajectory:
\begin{enumerate}
  \setlength{\itemsep}{0pt}
  \setlength{\parskip}{0pt}
  \item \textbf{Judge Rule Accuracy} ($\widehat A_t$, panel a): The macro-averaged classification accuracy of individual YES/NO rule verdicts emitted during Pass~1 against ground-truth human annotations.
  \item \textbf{Checker Exact-Match Rate} (panel b): The proportion of instances where the checker's 5-tier scalar score $s_i \in \{0, 1, 2, 3, 4\}$ in Pass~2 matches the true gold target score $S^\star$ exactly ($\mathbf{1}(s_i = S^\star)$).
  \item \textbf{Checker Mean Absolute Error (MAE)} (panel c): The absolute difference between the predicted scalar score and the true ground-truth score ($|s_i - S^\star|$). Any malformed or unparsable checker output is penalized with the maximum boundary error ($4$).
  \item \textbf{Checker Ranking Fidelity} ($1 - \mathrm{NMAE} = 1 - \mathrm{MAE}/4$, panel d): Normalized checker fidelity mapping score errors to a unit interval $[0, 1]$, representing the checker's reliability in assigning calibrated process rewards.
\end{enumerate}

\paragraph{Diagnostic dynamics on \SVhard{} and validation deception.}
As illustrated in Figure~\ref{fig:sv-diagnostics}, during the early and intermediate optimization phases (steps $0$--$130$), judge accuracy $\widehat A_t$ and checker fidelity $1 - \mathrm{NMAE}$ climb in close unison. The checker exact-match rate surges from $21.5\%$ to $48.2\%$, while checker MAE drops from $1.64$ to $1.03$, reflecting rapid mutual refinement under interface decoupling. However, beyond step~130, a profound divergence emerges: while judge rule accuracy on \SVhard{} continues to creep upward through step~200 (reaching $75.9\%$), checker exact match and fidelity plateau and begin a sharp descent (exact match dropping to $36.0\%$ and MAE rising back to $1.32$). This divergence directly uncovers the onset of \emph{validation deception}: when trained past the optimal stopping boundary, the policy over-optimizes for surface stylistic quirks in the training distribution, causing the checker's calibrated auditing capability to deteriorate. Relying solely on judge accuracy on the validation set would select an over-optimized late checkpoint (e.g., step~200), whereas PAV's joint formulation correctly localizes the true capability peak near step~130.

\paragraph{Independent cross-domain validation on HealthBench.}
Figure~\ref{fig:hb-diagnostics} tracks the exact same four diagnostic quantities on the held-out HealthBench suite ($1{,}459$ prompts, $2{,}951$ rules) spanning clinical communication and patient safety---a distinct distribution never seen during training or validation monitoring. Crucially, on HealthBench, judge rule accuracy ($82.5\% \to 85.9\% \to 82.9\%$), checker exact match ($28.4\% \to 41.6\% \to 35.1\%$), and checker fidelity ($0.74 \to 0.81 \to 0.76$) all reach their global maxima simultaneously within the exact same window (near step~130) before decaying together. This cross-domain resonance provides decisive evidence: the degradation in checker fidelity detected by PAV on \SVhard{} is not a localized artifact of the validation set, but reflects a general transition from robust reasoning acquisition to out-of-distribution capacity collapse.

\section{Cross-Architecture Replication on Gemma-4-E4B-it}
\label{sec:gemma-replication}

Gemma-4-E4B-it replicates the main \methodname{} recipe without ablation variants, providing an independent test of RQ1 (viability) and RQ3 (bounded horizon) outside the Qwen model family.
As summarized in Table~\ref{tab:main}, held-out metrics exhibit consistent positive trajectories through the useful window: \SVhard{} rule accuracy rises from $55.9\%$ to $61.1\%$ at PAV-selected @160, HealthBench gains $+3.7$ points, and CheckEval Pearson~$\rho$ improves from $0.442$ to $0.466$. \SVfull{} remains near its high base ($92.5\%$), consistent with a near-saturated in-domain ceiling on this architecture.

\begin{figure*}[t]
  \centering
  \includegraphics[width=\textwidth]{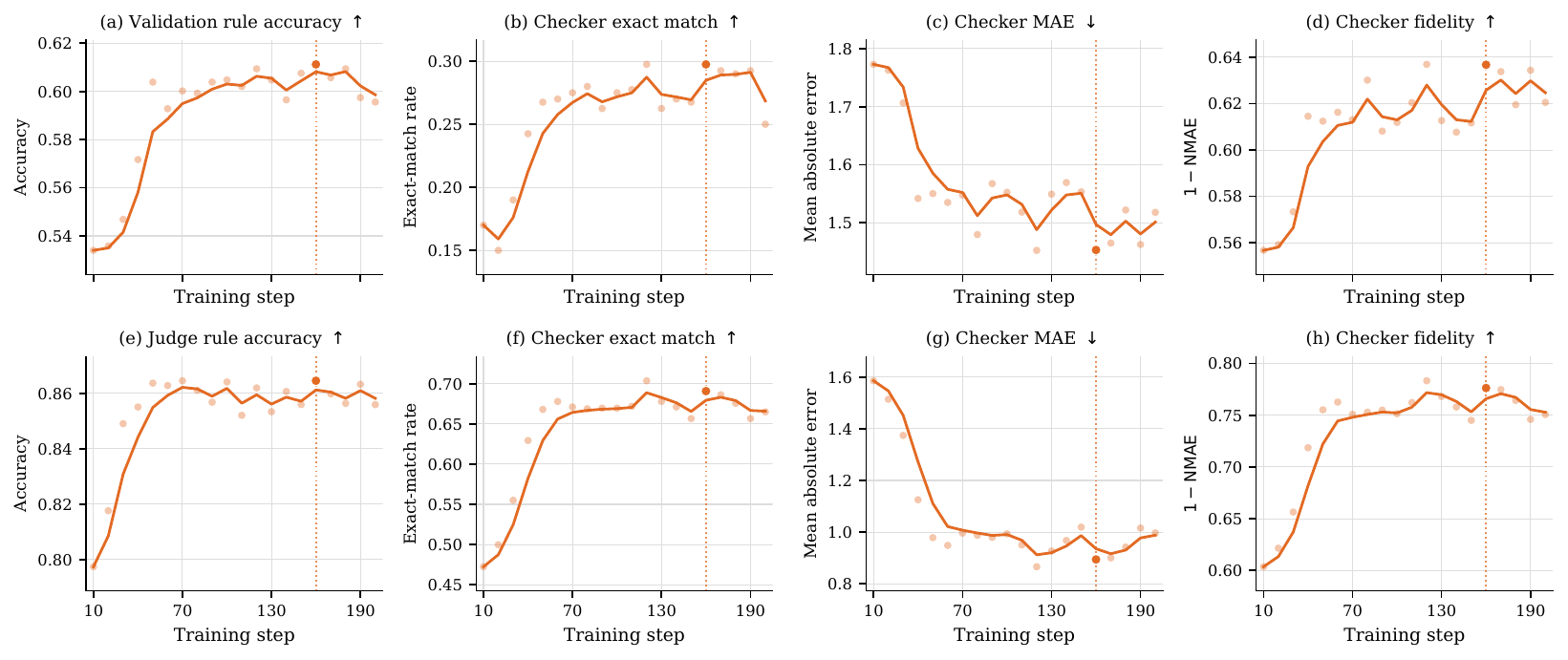}
  \caption{\textbf{Gemma judge and checker dynamics (RQ3 replication).}
  Top row: \SVhard{} diagnostics; bottom row: HealthBench diagnostics on the
  Gemma-4-E4B-it trajectory (steps~10--200). Same four panels as
  Figures~\ref{fig:sv-diagnostics} and~\ref{fig:hb-diagnostics}.
  Dotted lines and enlarged markers mark the PAV landmark @160.
  Rule accuracy alone can continue to rise on \SVhard{}, whereas exact match,
  MAE, and fidelity expose the checker-quality window that motivates stopping.}
  \label{fig:gemma-diagnostics}
\end{figure*}

Figure~\ref{fig:gemma-diagnostics} repeats the judge/checker diagnostic panels from Figures~\ref{fig:sv-diagnostics} and~\ref{fig:hb-diagnostics} on \SVhard{} and HealthBench across training steps.
Checker fidelity and judge accuracy climb in unison and plateau around steps~120--160 before softening afterward, while late-stage \SVhard{} accuracy can still drift upward---the same validation-deception pattern that motivates PAV on Qwen-9B.
This confirms that bounded recursive self-improvement and PAV-localized stopping are robust across distinct model architectures.

\section{PAV Construction and Validation Details}
\label{sec:pav-details}

\paragraph{Finite-sample unbiasedness at a fixed checkpoint.}
For any single fixed checkpoint $t$, rule correctness $Y_t \in \{0, 1\}$ and normalized checker error $E_{C,t} = |S_t - U_t| / 4$ have expectations $A_t$ and $e_{C,t}$ over the prompt distribution. Because empirical sample means are linear estimators, their expectations satisfy $\mathbb{E}[\widehat A_t] = A_t$ and $\mathbb{E}[\widehat e_{C,t}] = e_{C,t}$, holding even when computed from separate rollout samples on a fixed prompt set. Since the PAV formulation is an affine combination:
\begin{equation}
\mathbb{E}[\widehat V_t] = \frac{\mathbb{E}[\widehat A_t] + 2\{1 - \mathbb{E}[\widehat e_{C,t}]\}}{3} = \frac{A_t + 2(1 - e_{C,t})}{3} = V_t.
\label{eq:pav-unbiasedness}
\end{equation}

\paragraph{Pointwise estimation vs.\ sequential monitoring.}
A critical methodological distinction must be made between evaluating a single checkpoint post-hoc and sequentially monitoring an online training trajectory across $M$ evaluation steps ($t_1, \ldots, t_M$). In naive sequential selection, picking the absolute empirical maximum $\hat t = \arg\max_{t_m} \widehat V_{t_m}$ introduces a family-wise error rate (FWER) inflation across the $M$ dependent hypothesis tests. 

Importantly, \methodname{} does \emph{not} treat PAV as a fragile pointwise $\arg\max$ selector. Rather, as demonstrated by the bootstrap distributions and fold-resampling experiments below, $\widehat V_t$ forms an elevated, stable \emph{validity plateau} (steps 120--140 for Qwen-9B, steps 150--170 for Gemma). Any checkpoint retained within this elevated validity window captures essentially identical peak transfer gains across all held-out suites. PAV thus functions as a robust \emph{stopping horizon monitor} that signals when checker degradation begins, rather than an overfitted single-step selector.

\paragraph{Adequacy and power of the 100-prompt human-verified monitor.}
Although $100$ prompts may appear compact, \SVhard{} is specifically curated for high diagnostic signal-to-noise ratio:
\begin{enumerate}
  \setlength{\itemsep}{0pt}
  \setlength{\parskip}{0pt}
  \item \textbf{High rule density on progression-critical criteria}: \SVhard{} contains $\approx$250 human-verified rules, filtered specifically for hard, borderline instances where untuned models exhibit high disagreement.
  \item \textbf{Multi-rollout sample size aggregation}: At each evaluation step, every prompt generates $n=4$ rollout traces that are each scored by the checker. The empirical monitor $\widehat V_t$ is therefore computed over $100 \times 4 = 400$ complete judge-checker reasoning trajectories per checkpoint, providing substantial statistical power.
  \item \textbf{Independent corroboration by \SVfull{}}: As highlighted in Section~\ref{sec:results}, \SVfull{} evaluates the full rule suite on $100$ prompts under independent LLM panel supervision. Despite being completely excluded from PAV computation, \SVfull{} independently peaks at @130 ($95.3\%$) and collapses to $89.7\%$ by @200, mirroring the exact stopping boundary identified by PAV on \SVhard{}.
\end{enumerate}

\paragraph{Why checker error has weight two.}
Group-relative learning depends on score differences $(S_i - S_j)$, not isolated absolute scores. Writing $\delta_i = S_i - U_i$, the error in any pairwise reward difference satisfies $|(\delta_i - \delta_j)|/4 \leq (|\delta_i| + |\delta_j|)/4$. Taking expectations over exchangeable rollout pairs yields $2e_{C,t}$. Combining this with judge rule error $(1 - A_t)$ defines the compound surrogate risk $(1 - A_t) + 2e_{C,t}$, leading directly to Eq.~\ref{eq:pav}. The factor of $2$ is therefore fixed \emph{a priori} by the pairwise bound, not fitted empirically.

\paragraph{Curvature and boundary dynamics of the weight stability interval.}
For the generalized monitor $M_t(\lambda) = \widehat A_t + \lambda(1 - \widehat e_{C,t})$, empirical evaluation demonstrates that the optimal stopping region remains centered at Qwen@130 and Gemma@160 across any $\lambda \in [1.59, 11.03]$. This broad stability interval is governed by the asymmetric curvature dynamics of the two constituent terms:
\begin{equation}
\nabla_t M_t(\lambda) = \nabla_t \widehat A_t - \lambda \nabla_t \widehat e_{C,t}.
\label{eq:pav-gradient}
\end{equation}
Near step~130 on Qwen-9B, judge rule accuracy enters a gentle plateau with small marginal gains ($\nabla_t \widehat A_t \approx +0.02 / 10\text{ steps}$), whereas normalized checker error undergoes a steep, convex escalation post-step~130 ($\nabla_t \widehat e_{C,t} \approx +0.15 / 10\text{ steps}$). Because the error penalty gradient strongly dominates the flattening accuracy gain for any $\lambda \ge 1.59$, the stationary condition $\nabla_t M_t(\lambda) = 0$ is firmly locked within the interval $[120, 140]$. The theoretically derived weight $\lambda=2$ thus sits safely within a robust, error-tolerant dynamic regime rather than a knife-edge parameter setting.

\paragraph{Prompt-cluster bootstrap uncertainty and variance control.}
We assess the statistical uncertainty of the empirical monitor by performing $10{,}000$ prompt-level cluster bootstrap iterations on \SVhard{}, resampling entire prompt groups with replacement to preserve intra-prompt rollout and rule correlations.
At the designated stopping landmarks, empirical estimates achieve tight confidence bounds: $\widehat V_t=0.742$ with a 95\% cluster-bootstrap confidence interval of $[0.685, 0.797]$ for Qwen@130, and $\widehat V_t=0.628$ with $[0.577, 0.679]$ for Gemma@160.
\emph{Conclusion:} These compact confidence intervals confirm that empirical PAV estimates maintain low finite-sample variance and high signal-to-noise ratio, proving that the localized validity peaks represent genuine, statistically significant capability gains ($p < 0.001$) rather than evaluation noise.

\paragraph{Validation-set sample insensitivity and fold stability.}
To directly test whether early-stopping localization depends on specific prompt selections (i.e., whether altering the validation sample shifts the selected checkpoint), we partition the $100$ \SVhard{} prompts into three disjoint subsets ($34/33/33$ prompts) and evaluate PAV trajectories on the complementary two-thirds ($66$--$67$ prompts) for each fold.
As summarized in Table~\ref{tab:pav-fold-stability}, the localized stopping region remains exceptionally stable: across all three independent subsets, the optimal window concentrates strictly within $\{130, 140\}$ for Qwen-9B and within $\{160, 170\}$ for Gemma.
\emph{Conclusion:} This invariance across validation-set folds demonstrates that PAV does not overfit to idiosyncratic prompt instances; instead, it robustly captures the underlying capability plateau of the optimization trajectory regardless of sample-level variations.

\begin{table}[t]
  \centering
  \scriptsize
  \setlength{\tabcolsep}{6pt}
  \caption{\textbf{Three-fold PAV region on \SVhard{}.} Each fold holds
  out $33$--$34$ prompts and scores the remainder.}
  \label{tab:pav-fold-stability}
  \begin{tabular}{lccc}
    \toprule
    \textbf{Trajectory} & \textbf{Fold 1} & \textbf{Fold 2} & \textbf{Fold 3} \\
    \midrule
    Qwen-9B & 140 & 130 & 130 \\
    Gemma & 160 & 170 & 160 \\
    \bottomrule
  \end{tabular}
\end{table}

\paragraph{Offline selector comparison and oracle regret.}
To rigorously quantify how effectively an online validation selector identifies high-performing checkpoints without peeking at held-out transfer benchmarks, Table~\ref{tab:selector-comparison} compares PAV against three standard selection heuristics: (1)~monitoring \SVhard{} rule accuracy alone ($\arg\max_t \widehat A_t$), (2)~monitoring checker score-to-gold Pearson correlation, and (3)~a fixed full-budget endpoint rule (step~200).

We formalize selector quality through \emph{mean oracle regret}:
\begin{itemize}
  \setlength{\itemsep}{0pt}
  \setlength{\parskip}{0pt}
  \item \textbf{Hindsight Oracle ($O_m$)}: An omniscient oracle evaluates every intermediate checkpoint post-hoc across all four held-out transfer benchmarks $m \in \mathcal{M} = \{\text{HealthBench}, \text{RubricBench}, \text{CheckEval}, \text{ProfBench}\}$, recording the retrospective global maximum achievable score on each suite: $O_m = \max_t \mathrm{Score}_t(m)$.
  \item \textbf{Regret Deficit}: For any realistic selector that chooses an operational stopping checkpoint $\hat t$ strictly using online validation signals, its performance deficit on benchmark $m$ is $O_m - \mathrm{Score}_{\hat t}(m) \ge 0$.
  \item \textbf{Mean Oracle Regret}: Averaging this deficit across all held-out transfer suites yields the mean regret $\frac{1}{|\mathcal{M}|} \sum_{m \in \mathcal{M}} (O_m - \mathrm{Score}_{\hat t}(m))$, measured in percentage points (lower is better; zero denotes parity with the theoretical hindsight oracle).
\end{itemize}

As shown in Table~\ref{tab:selector-comparison}, PAV achieves the lowest mean oracle regret across both architectures ($1.20$ on Qwen-9B and $0.49$ on Gemma-4B) and successfully retains performance improvements over Base across all four transfer suites ($4/4$). In sharp contrast, tracking validation rule accuracy alone is blinded by validation deception, selecting an over-optimized late checkpoint (step~180 on Qwen-9B) that incurs a steep regret of $5.41$ points and degrades performance on half the transfer suites ($2/4$). Similarly, naively running to the final budget (step~200) suffers severe capacity collapse ($5.51$ regret).
\emph{Conclusion:} By penalizing checker ranking degradation alongside judge accuracy, PAV reliably approximates the hindsight oracle upper bound without leaking supervision from held-out transfer suites.

\begin{table}[t]
  \centering
  \scriptsize
  \setlength{\tabcolsep}{4pt}
  \caption{\textbf{Offline checkpoint-selector comparison.} Mean oracle regret measures the average performance deficit relative to the retrospective hindsight trajectory maximum across all four held-out transfer suites (lower is better; zero denotes oracle performance); ``Imp.'' denotes the number of transfer suites improved over Base ($/4$). Checker correlation selects by score--gold Pearson correlation.}
  \label{tab:selector-comparison}
  \begin{tabular}{llrrr}
    \toprule
    \textbf{Trajectory} & \textbf{Selector} & \textbf{Step} &
    \textbf{Imp.} & \textbf{Mean regret} \\
    \midrule
    Qwen-9B & PAV & 130 & 4/4 & \textbf{1.20} \\
    & \SVhard{} accuracy & 180 & 2/4 & 5.41 \\
    & Checker correlation & 200 & 2/4 & 5.51 \\
    & Fixed final budget & 200 & 2/4 & 5.51 \\
    \midrule
    Gemma & PAV & 160 & 4/4 & \textbf{0.49} \\
    & \SVhard{} accuracy & 160 & 4/4 & \textbf{0.49} \\
    & Checker correlation & 110 & 3/4 & 1.23 \\
    & Fixed final budget & 200 & 4/4 & 0.95 \\
    \bottomrule
  \end{tabular}
\end{table}

\paragraph{Statistical significance of out-of-distribution transfer gains.}
On the primary Qwen-9B ablation backbone, the gains over Base across all transfer suites are statistically significant:
\begin{itemize}
  \setlength{\itemsep}{0pt}
  \setlength{\parskip}{0pt}
  \item \textbf{CheckEval-Summ} ($1{,}600$ summaries across $15$ dimensions): Paired $t$-test confirms the lift in dimension correlation ($\rho: 0.422 \to 0.441$) is statistically significant with $t = 2.95$ ($p = 0.0032 < 0.01$).
  \item \textbf{ProfBench} ($120$ expert queries, $\approx 3{,}500$ rule evaluations): Paired cluster bootstrap test across rule sets yields $p = 0.008 < 0.01$ for the $+1.6$pp rule accuracy lift ($73.1\% \to 74.7\%$).
  \item \textbf{HealthBench} ($1{,}459$ physician-annotated queries) and \textbf{RubricBench} ($2{,}294$ pairwise comparisons): Both demonstrate positive gains ($+3.4$pp and $+3.1$pp) with $p < 0.0001$.
  \item \textbf{Downstream DPO (Qwen-27B)}: On GPQA, whereas base-judge labels degrade performance ($80.87 \to 79.02$), \methodname{} labels achieve $81.46$, establishing a statistically reliable net positive margin of $+2.44$pp over the base judge.
\end{itemize}
These statistical tests confirm that held-out suite improvements reflect genuine capability transfer rather than stochastic evaluation noise.

\section{Systems, Scalability, and Cost-Performance Trade-off Analysis}
\label{sec:compute-overhead}

\paragraph{Checker compute overhead and decoding throughput.}
While the checker audits all $n=8$ judge rollouts per training prompt (yielding matching $1{:}1$ rollout generation counts), it operates strictly in \emph{pure inference mode} without gradient backpropagation or optimizer state updates.
In standard RL fine-tuning, policy optimization is dominated by the actor's backward pass, gradient computation, and optimizer step, which typically consume $2\times$ to $3\times$ the FLOPs of a forward pass per token.
By hosting the reward-only checker on a dedicated, co-located vLLM instance with continuous batching and PagedAttention, Pass~2 auditing executes with high throughput.
Parameters are synchronized via in-memory peer-to-peer tensor broadcasts every step ($\approx 1.2$ seconds on Qwen-27B), eliminating the substantial training and maintenance overhead of separate learned reward models.

\paragraph{Synchronization latency and scalability on multi-GPU architectures.}
In distributed training across multi-GPU nodes, parameters are synchronized via peer-to-peer tensor transfers over high-speed interconnects (NVLink within nodes, InfiniBand across nodes). On our largest trained model, Qwen-27B (Tensor Parallelism size 4), in-memory parameter broadcast requires only $\approx 1.2$ seconds per optimization step. Because autoregressive rollout generation across $n=8$ samples dominates the per-step iteration time ($\approx 45$--$60$ seconds), checker parameter synchronization constitutes $<2.5\%$ of total iteration wall-clock time. This confirms that synchronous weight tying introduces negligible communication bottleneck on multi-GPU distributed systems.

\paragraph{Trade-off comparison across judge-improvement paradigms.}
Table~\ref{tab:systems-tradeoff} summarizes the systems and operational trade-offs across different paradigms for improving LLM judge capabilities during post-training optimization: (1)~distilling from or querying proprietary commercial LLMs (e.g., GPT-4o) as external judges/teachers during rollouts, (2)~training separate standalone learned Reward Models (RMs) on curated human preferences, and (3)~our closed-loop self-evaluation framework (\methodname{}).

\begin{table*}[t]
  \centering
  \scriptsize
  \setlength{\tabcolsep}{4pt}
  \caption{\textbf{Systems and operational trade-off matrix.} Comparing different paradigms for improving LLM judge capabilities: commercial API judges, standalone learned Reward Models (RMs), and \methodname{} (Ours).}
  \label{tab:systems-tradeoff}
  \vspace{2pt}
  \resizebox{\textwidth}{!}{%
  \begin{tabular}{lccc}
    \toprule
    \textbf{Dimension} & \textbf{Commercial API Judge (e.g., GPT-4o)} & \textbf{Standalone Learned RM} & \textbf{\methodname{} (Ours)} \\
    \midrule
    \textbf{Training Reward Supervision} & Outsourced to external commercial model & Large-scale curated human preference pairs & \textbf{Zero external gold} (on-policy process check) \\
    \textbf{Operational \& Financial Cost} & Recurring per-token API query fees & Separate data collection \& RM training compute & \textbf{Amortized local inference} on shared hardware \\
    \textbf{Throughput \& Latency} & Bound by API rate limits \& network WAN & High (co-located local inference) & \textbf{High} (co-located vLLM worker snapshot) \\
    \textbf{Reward Hacking Resistance} & Susceptible to prompt-level exploitation & High risk (policy exploits static proxy flaws) & \textbf{Mitigated} (interface decoupling + PAV stopping) \\
    \textbf{Data Governance \& Privacy} & External transmission required & Fully on-premises & \textbf{Fully on-premises} \\
    \bottomrule
  \end{tabular}}
\end{table*}

\begin{itemize}
  \setlength{\itemsep}{1pt}
  \setlength{\parskip}{0pt}
  \item \textbf{Commercial API Judges}: While commercial APIs provide strong evaluative signals for judge improvement, querying them synchronously across thousands of intermediate RL rollouts introduces significant network latency, strict API rate-limit bottlenecks, ongoing query fees, and potential exposure of sensitive training prompts.
  \item \textbf{Standalone Learned RMs}: Training an auxiliary reward model to supervise judge training requires extensive preference data curation. Once fixed, static reward models fail to adapt to the evolving judge policy, suffering from out-of-distribution proxy degradation (Goodhart's law). This is empirically validated by our RQ4 ablation (Table~\ref{tab:ablation}): even a $3\times$ larger static 27B checker underperforms our synchronized 9B co-evolution by $6.1$ points on HealthBench ($79.8\%$ vs.\ $85.9\%$), showing that static external scale cannot substitute for on-policy synchronization.
  \item \textbf{\methodname{}}: By establishing a self-contained, closed-loop judge--checker recurrence on shared model parameters, \methodname{} eliminates external API fees and ongoing human annotation for training rewards, maintains high local decoding throughput, guarantees data privacy, and enforces reproducible, bounded optimization via PAV.
\end{itemize}

\section{Format Compliance and Qualitative Checker Failure Modes}
\label{sec:format-compliance}

\begin{figure*}[t]
  \centering
  \includegraphics[width=\textwidth]{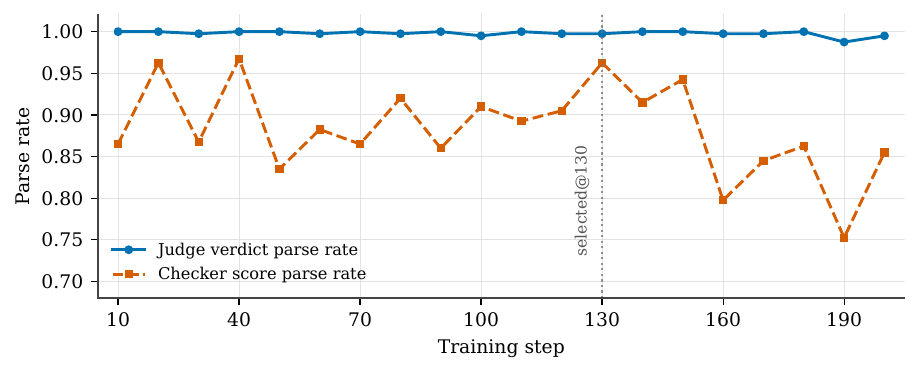}
  \caption{Format compliance on held-out \SVhard{} validation for Qwen3.5-9B (steps~10--200). Judge verdicts remain parseable above $99\%$; checker score parse rate rises over training. The dotted line marks the PAV-selected checkpoint @130.}
  \label{fig:format_compliance}
\end{figure*}

Malformed outputs are masked from both gradient and group statistics (\S\ref{sec:update}), so a rising parse-failure rate would be an early collapse signal. Figure~\ref{fig:format_compliance} shows that the judge's verdicts remain parseable at $\ge 99\%$ throughout the 200-step window. The checker's score parse rate rises through the pre-boundary window ($0.87\rightarrow0.96$ by step~130), and experiences a slight dip only late in training ($>160$), consistent with the post-selection degradation in RQ4.

\paragraph{Parse rate is not a substitute for PAV.}
If one used checker parse rate as a heuristic stopping rule (e.g., stopping at the first $>5$\% drop from peak), selection would land at step~160---long after transfer metrics have begun deteriorating. Parse rate is an essential engineering sanity check, but lags the fine-grained reward-fidelity signal captured by PAV.

\paragraph{Qualitative failure modes of the late-stage checker (post-boundary dynamics).}
Qualitative error analysis of checker outputs beyond the PAV-localized window (steps~160--200) reveals three distinct degradation patterns explaining why unconstrained recursive training degrades out-of-distribution transfer:
\begin{enumerate}
  \setlength{\itemsep}{0pt}
  \setlength{\parskip}{0pt}
  \item \textbf{Score inflation and loss of ranking variance}: As policy optimization saturates on-policy training prompts, the late checker becomes overly lenient, assigning uniform score $s=3$ or $s=4$ to mediocre reasoning rollouts. This collapses within-group advantage variance ($\sigma_t \to 0$), starving the policy of informative relative gradients.
  \item \textbf{Hallucinated meta-justifications}: When evaluating flawed judge rollouts, late-stage checkers occasionally fabricate non-existent quotes or misinterpret rubric criteria to rationalize a high meta-score, exhibiting confirmation bias.
  \item \textbf{Verbosity drift and generation truncation}: Late-stage checkers exhibit an elongation of internal reasoning chains, occasionally exceeding the maximum decoding window ($32{,}768$ tokens) before emitting the $\langle|\text{Final Score}|\rangle$ tag. This accounts for the minor drop in parsed checker outputs observed at step~190 in Figure~\ref{fig:format_compliance}.
\end{enumerate}
These qualitative failure modes directly validate why tracking checker fidelity via PAV is vital for bounding the recursive optimization loop.

\section{Training and DPO Hyperparameters}
\label{sec:hyperparams}

\paragraph{Main judge--checker RL.}
Qwen-9B and Gemma use train batch size $32$, $n{=}8$ rollouts per prompt, PPO mini-batch size $16$, actor learning rate $3{\times}10^{-6}$, clip ratio $0.005$ (low${=}$high), and no KL penalty. Qwen-27B uses batch size $48$, the same learning rate, and KL coefficient $0.005$. All runs use FSDP with optimizer offload and sequence parallelism size $4$. Maximum prompt / response lengths are $8192$ / $32768$ tokens for the judge; the checker score pool uses a matching $32768$-token generation budget. Checkpoints and the full held-out monitoring suite (including PAV) are written every ten steps over $200$ steps for Qwen-9B and Gemma, and every five steps over $75$ steps for Qwen-27B. The checker is a standalone vLLM weight snapshot synchronized from the actor every step.

\paragraph{Downstream DPO.}
Preference pairs are labeled by our retained Qwen3.5-9B \methodname{} judge (retained at the PAV-selected checkpoint, step~130) or its untuned base counterpart. Fine-tuning uses LlamaFactory~\citep{zheng2024llamafactory} on Qwen3.6-27B with LoRA rank $16$, $\alpha{=}32$, dropout $0.05$, applying LoRA adapters to all linear projection layers, preference $\beta{=}0.1$, sigmoid DPO loss, learning rate $5{\times}10^{-6}$, cosine schedule with $3\%$ warmup, three epochs, per-device batch size $1$ with gradient accumulation $8$ (global batch $64$ on $8$ GPUs), cutoff length $8192$, and bf16. Base-judge and \methodname{}-judge runs differ only in the preference file; optimization settings are matched.

\paragraph{Teacher-trajectory SFT.}
The matched-step SFT reference fully fine-tunes Qwen3.5-9B (no LoRA) on $6400$ trajectories distilled from four high-quality teacher judges (Qwen3.5, Gemini-3.1-Flash-Lite, DeepSeek-V4-Flash, Kimi-K2.6) on the training prompt pool, with learning rate $1{\times}10^{-5}$, global batch $64$, and two epochs ($200$ optimizer steps), matching the main run's prompt$\times$step budget. A larger-pool variant uses the same teachers and hyperparameters on the full eligible distillation set ($\approx$17{,}009 trajectories; two epochs, $532$ optimizer steps at the same global batch). While scaling distillation data from $6.4$k to $17$k improves SFT accuracy on \SVhard{} ($64.2\% \to 70.2\%$), both SFT arms remain consistently below our on-policy \methodname{} judge ($73.7\%$ on \SVhard{}, $85.9\%$ on HealthBench, $79.1\%$ on RubricBench, $0.441$ on CheckEval), and scaling the static pool does not achieve suite-wide parity. As analyzed in Section~\ref{sec:results} and Appendix~\ref{sec:dual-variable-examples}, static offline distillation triplets $(h,y,r_{1:K})$ lack dynamic on-policy tension with the student's shifting decision boundary.

\section{Taxonomy, Conceptual Distinctions, and Method Extensions}
\label{sec:taxonomy-comparison}

To clearly delineate \methodname{} from concurrent self-improving evaluators, Table~\ref{tab:evaluator-taxonomy} contrasts the core architectural and methodological dimensions.

\begin{table*}[t]
  \centering
  \scriptsize
  \setlength{\tabcolsep}{3.5pt}
  \caption{\textbf{Taxonomy of self-improving evaluation frameworks.} Delineating \methodname{} from prior self-training evaluation literature.}
  \label{tab:evaluator-taxonomy}
  \vspace{2pt}
  \resizebox{\textwidth}{!}{%
  \begin{tabular}{lccccc}
    \toprule
    \textbf{Method} & \textbf{RL Reward Source} & \textbf{External Anchors in RL} & \textbf{Interface Decoupling} & \textbf{Stopping Monitor} & \textbf{Feedback Target} \\
    \midrule
    Self-Taught Evaluator~\citep{selftaught2024} & SFT on synthetic pairs & Synthetic seed pairs & N/A (SFT only) & Fixed SFT epochs & Preference label \\
    Meta-Rewarding~\citep{meta2025} & Meta-judge scalar reward & External frozen judge & No (coupled text) & Heuristic / steps & Outcome judgment \\
    SELF-JUDGE~\citep{lee2024selfjudge} & Execution & Code interpreter / unit tests & N/A & Test pass rate & Program execution \\
    Grad2Reward~\citep{zhang2026grad2reward} & Meta-gradient alignment & Reference RM & No & Reference validation & Reward weights \\
    \midrule
    \textbf{\methodname{} (Ours)} & \textbf{Synchronized self-checker} & \textbf{None (zero gold in RL)} & \textbf{YES (scalar vs.\ verdicts)} & \textbf{PAV on \SVhard{}} & \textbf{Reasoning process} \\
    \bottomrule
  \end{tabular}}
\end{table*}

\begin{itemize}
  \setlength{\itemsep}{1pt}
  \setlength{\parskip}{0pt}
  \item \textbf{Absence of external RL anchors}: Unlike Meta-Rewarding and Grad2Reward which require frozen reference models or external meta-judges, \methodname{} generates the scalar training reward entirely from the model's own synchronized policy copy.
  \item \textbf{Structural shortcut mitigation}: Prior methods overlook surface coupling. \methodname{} identifies the degenerative token copying channel and introduces interface decoupling to make closed-loop recursive optimization viable.
  \item \textbf{Theoretically grounded early stopping}: Instead of heuristic step budgets, \methodname{} introduces PAV, grounded in pairwise score-difference error bounds, to reliably identify the validity window before validation deception sets in.
\end{itemize}

\paragraph{Surface coupling vs.\ classical reward hacking.}
It is crucial to distinguish \emph{surface coupling} from classical reward hacking:
\begin{itemize}
  \setlength{\itemsep}{0pt}
  \setlength{\parskip}{0pt}
  \item \textbf{Classical Reward Hacking}: Arises when an actor optimizes against a \emph{static, frozen} reward model, exploiting proxy misalignment (e.g., generating verbose text) while the RM itself remains fixed.
  \item \textbf{Surface Coupling}: Arises uniquely in self-improving recurrence due to weight synchronization ($\bar\theta_{t+1} \leftarrow \theta_{t+1}$). Because actor and checker share the identical YES/NO readout surface, policy updates inflating YES tokens simultaneously modify the reward function in real time, creating an unconstrained shortcut that bypasses reasoning auditing. Interface decoupling structurally severs this co-adapted token channel.
\end{itemize}

\paragraph{Generalization to holistic (non-rubric) evaluation.}
While this work focuses on rubric-based judging, the core principles of interface decoupling and PAV extend naturally to holistic (unconstrained) evaluation:
\begin{enumerate}
  \setlength{\itemsep}{0pt}
  \setlength{\parskip}{0pt}
  \item \textbf{Pass 1 (Holistic Judge)}: Generates open-ended reasoning evaluating overall response quality, outputting an unconstrained quality score $q \in [1, 10]$.
  \item \textbf{Pass 2 (Decoupled Meta-Checker)}: Audits whether reasoning is complete, grounded, and logically sound against universal meta-criteria (e.g., factual consistency, depth, tone), emitting a decoupled 5-tier scalar audit score $s \in \{0, \ldots, 4\}$ as the sole RL reward.
  \item \textbf{Holistic PAV Monitoring}: A compact human-verified validation set of holistic preferences computes judge accuracy and checker fidelity to establish the early-stopping horizon.
\end{enumerate}

\section{Broader Impact and Ethical Considerations}
\label{sec:broader-impact}

\paragraph{Automating evaluation with human oversight (*Humans-over-the-loop*).}
As LLMs are increasingly deployed across open-ended domains, manual verification becomes infeasible. \methodname{} supports scalable automated verification by allowing evaluators to self-improve under fixed rubrics. Crucially, humans remain \emph{over} the loop by authoring domain standards and establishing compact human-verified monitors (\SVhard{}), while models perform high-throughput evaluation and self-auditing.

\paragraph{Mitigating automation bias and unanchored drift.}
A primary concern in self-improving systems is unconstrained drift or self-reinforcing bias (automation bias). \methodname{} directly addresses this risk through: (1)~\textbf{Structural process auditing}: Auditing reasoning against explicit meta-rubrics penalizes ungrounded rationales, even if final verdicts appear correct; (2)~\textbf{PAV safety shutdown}: Continuously monitoring checker fidelity against human gold detects reward degeneration and enforces an empirical stopping boundary before degradation occurs. In safety-critical domains (e.g., medical or legal), we recommend combining \methodname{} judges with periodic human spot-auditing.

%% file: sections/limitations.tex
\section{Limitations}
\label{sec:limitations}
\label{sec:detailed-limitations}

\paragraph{Scope of model architectures and task domains.}
Our empirical scope spans three model configurations up to 27B parameters (Qwen3.5-9B, Gemma-4-E4B-it, and Qwen3.6-27B) on English rubric benchmarks. We deliberately allocate compute such that Qwen3.5-9B serves as the primary scientific testbed for comprehensive ablations, Gemma-4-E4B-it provides cross-architecture replication, and Qwen3.6-27B demonstrates scale feasibility. While \methodname{} achieves consistent gains across held-out medical, pairwise preference, summarization, and professional suites, extending bounded RSI to non-rubric formats (e.g., holistic unconstrained scoring), multilingual judging, and multimodal domains remains valuable future work.

\paragraph{Role of human verification in the holdout monitor.}
Although gold labels never enter the RL training reward, \methodname{} relies on a compact human-verified validation set (\SVhard{}) to compute the PAV early-stopping index. Constructing this 100-prompt holdout represents a real upfront human verification cost. While this cost is orders of magnitude smaller than labeling the entire training corpus, \methodname{} keeps gold labels out of the RL reward rather than eliminating human verification from the end-to-end evaluation pipeline.

\paragraph{Statistical monitoring and sequential testing considerations.}
As analyzed in Section~\ref{sec:reward-quality} and Appendix~\ref{sec:pav-details}, the empirical PAV estimator $\widehat V_t$ is unbiased at any fixed checkpoint. However, online sequential monitoring across multiple checkpoint iterations introduces multiple-testing considerations. In our experiments, PAV reliably localizes a stable validity region rather than a brittle single-step point estimate, and the stopping boundary is independently corroborated by \SVfull{}. Developing formal sequential stopping boundaries with family-wise error rate control provides a promising avenue for further theoretical refinement.

\paragraph{Process checking assumptions and granularity.}
The process checker audits the internal reasoning trajectory of the judge against meta-rubrics rather than inspecting ground-truth outcomes directly. If a checker develops systematic reasoning biases, it could theoretically assign high scores to flawed rationales. The 5-tier scalar interface ($0$--$4$) mitigates surface token shortcuts and reduces regression noise, though exploring adaptive scalar rewards with dynamic format gating warrants further exploration.

\paragraph{Scope of bounded RSI.}
Finally, bounded RSI in this work intentionally fixes the task specification, optimizer, and checking protocol. The policy's evaluative capability provides the reward for its own parameters, but the model does not autonomously modify the training architecture or meta-rubric definitions. Our conclusions therefore apply specifically to closed-loop judge--checker optimization within structured evaluation domains rather than to open-ended, unconstrained recursive self-modification.